\documentclass[acmsmall]{acmart}

\usepackage{amsmath}
\usepackage{paralist}
\usepackage{subfigure}
\usepackage{algorithm}
\usepackage{algorithmicx}
\usepackage[noend]{algpseudocode}

\AtBeginDocument{%
  }

\setcopyright{acmlicensed} \copyrightyear{2024} \acmYear{2024}
\acmDOI{XXXXXXX.XXXXXXX}

\acmJournal{PACMMOD} \acmVolume{0} \acmNumber{0} \acmArticle{0}
\acmMonth{0}

\begin{document}

\title{Online Detection of Anomalies in Temporal Knowledge Graphs with Interpretability}

\author{Jiasheng Zhang}
\affiliation{%
  \institution{School of Computer Science and Engineering, University of Electronic Science and Technology of China}
  \city{Chengdu}
  \country{China}}
\email{zjss12358@std.uestc.edu.cn} \orcid{0000-0002-5640-2020}

\author{Rex Ying}
\affiliation{%
  \institution{Department of Computer Science, Yale University}
  \city{New Haven}
  \state{Connecticut}
  \country{USA}}
\email{rex.ying@yale.edu} \orcid{0000-0002-5856-5229}

\author{Jie Shao}
\authornote{Corresponding author.}
\affiliation{%
  \institution{Shenzhen Institute for Advanced Study, University of Electronic Science and Technology of China}
  \city{Shenzhen}
  \country{China}}
\email{shaojie@uestc.edu.cn} \orcid{0000-0003-2615-1555}

\renewcommand{\shortauthors}{Zhang et al.}

\begin{abstract}
Temporal knowledge graphs (TKGs) are valuable resources for
capturing evolving relationships among entities, yet they are often
plagued by noise, necessitating robust anomaly detection mechanisms.
Existing dynamic graph anomaly detection approaches struggle to
capture the rich semantics introduced by node and edge categories
within TKGs, while TKG embedding methods lack interpretability,
undermining the credibility of anomaly detection. Moreover, these
methods falter in adapting to pattern changes and semantic drifts
resulting from knowledge updates. To tackle these challenges, we
introduce \textsc{AnoT}, an efficient TKG summarization method
tailored for interpretable online anomaly detection in TKGs.
\textsc{AnoT} begins by summarizing a TKG into a novel rule graph,
enabling flexible inference of complex patterns in TKGs. When new
knowledge emerges, \textsc{AnoT} maps it onto a node in the rule
graph and traverses the rule graph recursively to derive the anomaly
score of the knowledge. The traversal yields reachable nodes that
furnish interpretable evidence for the validity or the anomalous of
the new knowledge. Overall, \textsc{AnoT} embodies a
detector-updater-monitor architecture, encompassing a detector for
offline TKG summarization and online scoring, an updater for
real-time rule graph updates based on emerging knowledge, and a
monitor for estimating the approximation error of the rule graph.
Experimental results on four real-world datasets demonstrate that
\textsc{AnoT} surpasses existing methods significantly in terms of
accuracy and interoperability. All of the raw datasets and the
implementation of \textsc{AnoT} are provided in
\url{https://github.com/zjs123/ANoT}.
\end{abstract}

\ccsdesc[500]{Computing methodologies~Anomaly detection}
\ccsdesc[500]{Computing methodologies~Temporal reasoning}

\keywords{Temporal knowledge graph, Anomaly detection, Graph
summarization}


\maketitle

\section{Introduction}

Many human activities, such as political interactions
\cite{DBLP:journals/pvldb/FanJLTX22} and e-commerce
\cite{DBLP:journals/tkde/LiuZSOJYN23}, can be effectively
represented as time-evolving graphs with semantics, which are
referred to as temporal knowledge graphs (TKGs). These TKGs are
dynamic directed graphs characterized by node and edge categories.
In this context, nodes represent entities in the real world (e.g.,
\textit{United States}), while labeled edges signify the relations
between these entities (e.g., \textit{Born In}). Each edge in
conjunction with its connected nodes can constitute a tuple $(s, r,
o, t)$ that encapsulates a piece of real-world knowledge, where $s$
and $o$ denote the subject and object entities, $r$ denotes the
relation, and $t$ represents the occurrence timestamp of the
knowledge.

\begin{figure}[t]
  \centering
  \includegraphics[width=1.0\linewidth]{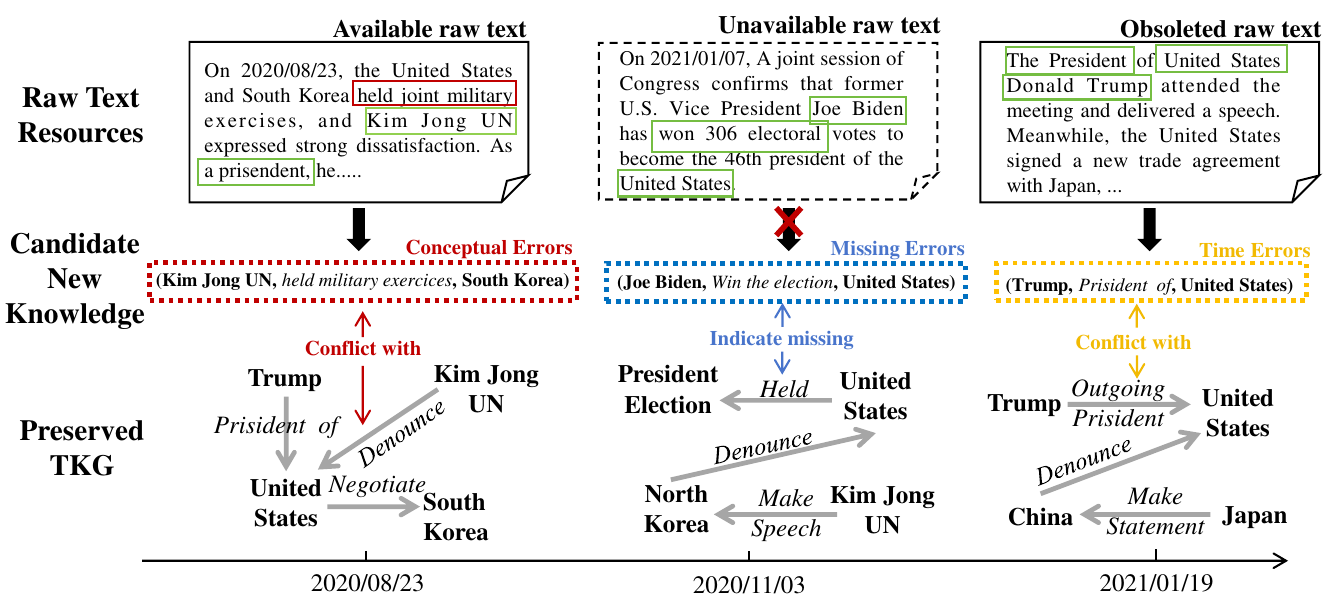}
  \caption{An illustration of different types of anomalies in TKGs and how anomalies relate to preserved knowledge.}
  \label{fig:anomalies}
\end{figure}

While TKGs have demonstrated their value across various applications
\cite{DBLP:journals/pvldb/ZhengYZC18,
DBLP:journals/pvldb/GuptaSGGZLL14}, they often contain numerous
anomalies that can significantly impede their reliability. As
illustrated in Figure~\ref{fig:anomalies}, TKG construction relies
on automatic extraction from unstructured text
\cite{DBLP:conf/icde/ZhuWXWJWMZ23}. However, existing extraction
techniques often encounter conceptual confusion
\cite{DBLP:journals/csur/NasarJM21} and inaccurate relation matching
\cite{DBLP:journals/csur/SmirnovaC19}, potentially introducing noisy
tuples with erroneous entities or relations, termed as
\textbf{conceptual errors}. Furthermore, ongoing interactions in the
real world lead to the formation of new knowledge or render existing
knowledge obsolete. However, knowledge-updating processes are often
insufficient and delayed \cite{DBLP:conf/emnlp/TangFZ19}, resulting
in either the omission of new knowledge or the retention of invalid
knowledge, termed \textbf{missing errors} and \textbf{time errors}.
Specifically, these errors are indicated by their conflicts with
preserved knowledge. Conceptual errors conflict with the interaction
preference of entities, while time errors conflict with other timely
updated knowledge in their time order. The missing errors are valid
knowledge not included in TKGs and should have few conflicts. TKG
anomaly detection refers to detecting conceptual, time, and missing
errors by measuring conflicts. Unfortunately, this field receives
little attention.

One closely related research field is dynamic graph anomaly
detection \cite{DBLP:journals/sigkdd/SalehiR18}, aimed at
identifying abnormal connections in time-evolving graphs. However,
existing methods primarily rely on simple structural properties,
such as connectivity \cite{DBLP:conf/icde/AggarwalZY11} or clusters
\cite{DBLP:conf/kdd/ManzoorMA16}, thus failing to capture the
intricate patterns present in TKGs such as relational closures
\cite{DBLP:journals/kbs/ZhangLSS22} and temporal paths
\cite{DBLP:journals/pvldb/FanJLTX22}. Furthermore, these methods do
not consider the semantic attributes of nodes and are thus
ineffective in handling the rich semantics introduced by node and
edge categories. Another related field is TKG embedding
\cite{DBLP:journals/corr/abs-2308-02457}, which aims to represent
entities and relations as low-dimensional vectors for downstream
tasks such as link prediction \cite{DBLP:journals/pvldb/RossiFMT22}.
While these vectors can also be utilized for anomaly scoring
\cite{DBLP:conf/www/JiaXCWE19}, they face limitations due to the
requirement of substantial training data
\cite{DBLP:journals/pvldb/GaoLXSDY19} and the absence of
interpretability \cite{DBLP:conf/kdd/JungJK21}, making their anomaly
detection less convincing. Moreover, the semantics of entities and
graph patterns change as TKGs continue to update. Since these
methods learn fixed vectors, they struggle to adapt to online
changes. In summary, TKG anomaly detection encounters three primary
challenges: \textbf{CH1.} Be interpretable to provide evidence for
the detection results; \textbf{CH2.} Capturing complex patterns
arising from entity and relation semantics and temporal relevance;
\textbf{CH3.} Handling semantic and pattern changes caused by
knowledge updates.

In this paper, we propose \textsc{AnoT}, a novel TKG summarization
method for online detection of anomalies. It distills TKG into an
interpretable rule graph and walks on it to gather evidence of
knowledge validity. For CH1, \textsc{AnoT} conceptualizes observed
knowledge as atomic rules, distilling interaction patterns within
TKGs into concept triples that are both concise and human-readable.
By mapping new knowledge as atomic rules, they can interpret how new
knowledge complies or conflicts with existing patterns, and thus
give evidence of the decision. Moreover, by identifying these
discrepancies, \textsc{AnoT} can offer guidelines on correcting
anomalous knowledge. For CH2, \textsc{AnoT} advances the utility of
atomic rules by associating them through two common occurring
relationships: chain and triadic occurrence. We formulate this
association results as a rule graph, with nodes representing
interaction patterns and directed edges indicating their sequential
relationships. This novel structure enables the flexible inference
of complex patterns by walking on it. For CH3, given the rule graph,
the semantic and pattern changes can be easily reformed as editions
on its nodes and edges, which provides a precise and scalable method
for uniformly managing online changes. Additionally, we propose an
error approximation strategy to determine the refresh time of the
rule graph. Conceptually, \textsc{AnoT} is composed of three
elements: a detector that constructs the rule graph and evaluates
new knowledge, an updater that revises the rule graph based on new
knowledge, and a monitor that estimates the rule graph's
availability. Based on the rule graph, we propose \textbf{static
scores} (measuring interaction preference conflicts) to detect
conceptual errors and \textbf{temporal scores} (measuring occurrence
order conflicts) to detect time errors. Consequently, the missing
errors can be filtered as knowledge with low static and time scores
during walking on the rule graph. Our contributions are as follows:
\begin{compactitem}
    \item To the best of our knowledge, we are the first to study anomaly
detection for TKGs. We propose an accurate and interpretable
solution \textsc{AnoT} for this problem.
    \item We are the first to study the method and application of TKG
summarization. We propose the rule graph to effectively summarize
TKGs, and verify its usage in anomaly detection.
    \item Extensive experiments on four real-world TKGs justify that
\textsc{ANoT} can accurately detect anomalies with efficiency while
being interpretable. It outperforms existing methods by an average
of 11.5\% in AUC and 13.6\% in precision.
\end{compactitem}

\section{Related Work}

\paragraph{\textbf{Dynamic graph anomaly detection.}}
Existing methods fall into two categories. One is statistical
methods, which leverage the shallow mechanisms to extract the
structural information \cite{DBLP:conf/icdm/EswaranF18,
DBLP:conf/sdm/RanshousHSS16, DBLP:conf/kdd/ManzoorMA16,
DBLP:conf/kdd/EswaranFGM18}. For example, CAD
\cite{DBLP:conf/sigmod/SricharanD14} detects abnormal edges by
tracking changes in structure and weight. DynAnom
\cite{DBLP:conf/kdd/GuoZS22} uses the dynamic forward push algorithm
to calculate the personalized PageRank vector for each node.
AnoGraph \cite{DBLP:conf/kdd/0001WKSYH23} extends the count-min
sketch data structure to detect anomalous edges through dense
subgraph searches. F-FADE \cite{DBLP:conf/wsdm/Chang0SASL21} models
the time-evolving distributions of node interactions using frequency
factorization. However, they cannot capture complex patterns brought
by entity and relation semantics in TKGs. The other is deep
learning-based methods, which detect anomalies by learning vector
representations for nodes \cite{DBLP:journals/tkde/LiuPWXWCL23,
DBLP:conf/icdm/FanYPZZXSXS020, DBLP:journals/eswa/BelleDTW22,
DBLP:conf/ijcai/ZhengLLLG19}. AEGIS \cite{DBLP:conf/ijcai/DingLAL20}
proposes a graph differentiation network to learn node
representations. Netwalk \cite{DBLP:conf/kdd/YuCAZCW18} combines
random walk and a dynamic clustering-based model to score anomalies.
AER \cite{DBLP:journals/pvldb/FangFGFH23} uses an anonymous
representation strategy to identify edges by their local structures.
TADDY \cite{DBLP:journals/tkde/LiuPWXWCL23} uses a dynamic graph
transformer model to aggregate spatial and temporal information.
However, they lack enough high-quality labels
\cite{DBLP:journals/pvldb/GaoLXSDY19} and the learned
representations are not interpretable, making their detection less
convincing.

\paragraph{\textbf{TKG embedding.}}
Factorization-based methods \cite{DBLP:journals/isci/ZhangCSCL24,
DBLP:journals/eswa/YangYSX24, DBLP:journals/corr/abs-2308-02457}
regard TKGs as 4-order tensors and use tensor factorization for
embeddings. TNT \cite{DBLP:conf/iclr/LacroixOU20}  builds on the
complex vector model ComplEx
\cite{DBLP:journals/jmlr/TrouillonDGWRB17} with temporal
regularization. Timeplex \cite{DBLP:conf/emnlp/JainRMC20} extends
TNT by capturing the recurrent nature of relations. TELM
\cite{DBLP:conf/naacl/XuCNL21} learns multi-vector representations
with canonical decomposition. However, they learn tensors with fixed
shapes, limiting their ability to handle new entities and
timestamps. Diachronic embedding-based methods
\cite{DBLP:conf/emnlp/DasguptaRT18, DBLP:conf/coling/XuNAYL20} model
entity representations as time-related functions. DE-simple
\cite{DBLP:conf/aaai/GoelKBP20} uses nonlinear operations to model
various evolution trends of entity semantics. ATiSE
\cite{DBLP:conf/semweb/XuNAYL20} uses multi-dimensional Gaussian
distributions to model the uncertainty of entity semantics.
TA-DistMult \cite{DBLP:conf/emnlp/Garcia-DuranDN18} uses a sequence
model for time-specific relation representations. However, they
over-simplify the evolution of TKG and ignore the graph structure.
GNN-based methods \cite{DBLP:conf/wsdm/ParkLMCFD22,
DBLP:conf/www/RenBXM23, DBLP:conf/icde/Liu0X0023} employ the
message-passing mechanism to simulate the entity interactions. TeMP
\cite{DBLP:conf/emnlp/WuCCH20} uses self-attention to model the
spatial and temporal locality. RE-GCN
\cite{DBLP:conf/sigir/LiJLGGSWC21} auto-regressively models
historical sequence and imposes attribute constraints on entity
representations. However, they lack interpretability and cannot
handle online changes.

\paragraph{\textbf{Graph summarization.}}
Graph summarization is closely related to anomaly detection since it
aims to find general patterns in data, and thus in turn can reveal
anomalies \cite{DBLP:conf/www/BelthZVK20,
DBLP:journals/vldb/CebiricGKKMTZ19}. Many studies on knowledge graph
summarization have focused on query-related summaries
\cite{DBLP:journals/pvldb/WuYSIY13} and personalized summaries
\cite{DBLP:conf/icdm/SafaviBFMMK19}, while KGist
\cite{DBLP:conf/www/BelthZVK20} learns inductive summaries by
introducing root graphs. However, they ignore the temporality of
knowledge. Recent efforts have focused on dynamic graph
summarization \cite{DBLP:conf/bigdataconf/HajiabadiST22}. TimeCrunch
\cite{DBLP:conf/kdd/ShahKZGF15} uses temporal phrases to describe
the temporal connectivity. PENminer \cite{DBLP:conf/kdd/BelthZK20}
mines activity snippets' persistence in evolving networks. However,
they only focus on evolving connectivity and fail to handle rich
semantics in TKGs. In this paper, we propose \textsc{AnoT}, a
scalable and information-theoretic method for inductive TKG
summarization.

\section{Preliminaries}

\subsection{Temporal Knowledge Graph}

A temporal knowledge graph is denoted as $\mathcal{G} =
(\mathcal{E}, \mathcal{R}, \mathcal{T}, \mathcal{F})$. $\mathcal{E}$
and $\mathcal{R}$ are entity set and relation set, respectively.
$\mathcal{T}$ is the set of observed timestamps and $\mathcal{F}$ is
the set of facts. In real-world scenarios, $\mathcal{E}$,
$\mathcal{T}$, and $\mathcal{F}$ will be continuously enriched. Each
tuple $(s, r, o, t) \in \mathcal{F}$ connects the subject and object
entities $s, o \in \mathcal{E}$ via a relation $r \in \mathcal{R}$
in timestamp $t \in \mathcal{T}$, which means a unit knowledge
(i.e., a fact). We represent the connectivity of $\mathcal{G}$ in
each timestamp $t$ with a $|\mathcal{E}| \times |\mathcal{E}| \times
|\mathcal{R}|$ adjacency tensor $A_t$, where 1 represents that the
entities are connected by the relation in timestamp $t$. There are
two common \textbf{occurring relationships} exist in TKGs. One is
\textbf{chain occurring} defined as $\{ (s,r_i,o,t_i) \rightarrow
(s,r_j,o,t_j) | t_j \geq t_i \}$, e.g., $(Obama, Win the Selection,
United States, 2008/11/04) \rightarrow (Obama, President of, United
States, 2009/01/20)$. The other is \textbf{triadic occurring}
defined as $\{ ((s,r_i,$ $o,t_i), (s,r_j,p,t_j)) \rightarrow
(o,r_k,p,t_k) | t_k \geq max(t_i, t_j)\}$, e.g., $((China, Host
Visit, Saudi Arabia, 2023/03/$ $06), (China, Host Visit, Iran,
2023/03/06) \rightarrow (Saudi Arabia, SignAgreement, Iran,
2023/03/10)$. The facts on the left of the arrow are called head
facts, and those on the right are called tail facts.

\subsection{Anomalies in TKGs}

Here, we formally define three kinds of typical anomalies in TKGs.
Given a TKG $\mathcal{G}$, we first define its corresponding ideal
TKG as $\hat{\mathcal{G}} = (\mathcal{E}, \mathcal{R}, \mathcal{T},
\hat{\mathcal{F}})$ which removes all the incorrect facts from
$\mathcal{F}$ and complete all the missing facts into $\mathcal{F}$
(i.e., $(s, r, o, t) \in \hat{\mathcal{F}}$ if and only if it holds
in reality). We further define the ideal triple set
$\hat{\mathcal{L}} = \{(s,r,o)|(s,r,o,t) \in \hat{\mathcal{F}}\}$.
Note that, $\hat{\mathcal{G}}$ and $\hat{\mathcal{L}}$ are only
conceptual aids that do not exist. We then use it to define
anomalies.

\subsubsection{Conceptual Errors.}

Extraction methods may introduce noised facts with error entities or
relations in TKGs. Formally, we define the conceptual errors as
$\mathcal{F}_c = \{ (s_c, r_c, o_c, t_c) | (s_c, r_c, o_c, t_c) \in
\mathcal{F}, (s_c, r_c, o_c) \notin \hat{\mathcal{L}} \}$, e.g.,
$(Joe Biden, Born In, Ireland, 1942/11/20)$.

\subsubsection{Time Errors.}

Knowledge updating may make existing facts invalid, but update
delays will let these invalid facts not be removed from TKGs.
Formally, we define the time errors as $\mathcal{F}_t = \{ (s_t,
r_t, o_t, t_t) | (s_t, r_t, o_t, t_t) \in \mathcal{F}, (s_t, r_t,
o_t) \in \hat{\mathcal{L}}, (s_t, r_t, o_t, t_t) \notin
\hat{\mathcal{F}} \}$. For example, $(Obama, President$ $of, United
States, 2023/10/21)$.

\subsubsection{Missing Errors.}

Insufficient updates also prevent some correct facts not being added
to TKGs. Formally, we define the missing errors as $\mathcal{F}_m =
\{ (s_m, r_m, o_m, t_m) | (s_m, r_m, o_m, t_m) \notin \mathcal{F},
(s_m, r_m, o_m,$ $t_m) \in \hat{\mathcal{F}} \}$. For instance, a
TKG might include the knowledge Barack Obama left office but lack
his inauguration. Unlike TKG completion that predicts missing
entities or relations for given tuples, we aim to find which tuple
is missing in TKG. Note that these anomalies will persist as TKGs
keep growing in real-world situations.

\subsection{Minimum Description Length Principle}

In the two-part minimum description length (MDL) principle
\cite{DBLP:journals/automatica/Rissanen78}, given a set of models
$\mathcal{M}$, the best model $M \in \mathcal{M}$ on data $D$
minimizes $L(M) + L(D | M)$, where $L(M)$ is the length (in bits) of
the description of $M$, and $L(D | M)$ is the length of the
description of the data when encoded using $M$. In this work, we
leverage MDL to find the optimal summarization model of a given TKG.
Each MDL-based approach must devise its definitions for the
description lengths, and here we follow the most commonly used
primitives \cite{DBLP:journals/datamine/Galbrun22}.

\subsection{Summarization of A TKG}

The summarization of a graph is a more refined and compact
representation of the graph
\cite{DBLP:journals/corr/abs-2203-05919}, including super-graphs
\cite{DBLP:journals/pvldb/ChenLFCYH09}, sparsified graphs
\cite{DBLP:conf/kdd/MaccioniA16}, and independent rules
\cite{DBLP:conf/kdd/ShahKZGF15}. However, they fail to handle rich
semantics and temporal relevance in TKGs, inspiring us to propose a
novel rule graph as the summarization of a TKG.

\subsubsection{Atomic Rules.}

Given a TKG $\mathcal{G}$, we first construct a function
$\mathcal{C}(\cdot)$, which takes each entity as input and outputs
its category. Based on this, each knowledge $(s, r, o, t) \in
\mathcal{G}$ can be mapped as an atomic rule $(\mathcal{C}(s), r,
\mathcal{C}(o))$, which summarizes the interaction pattern of the
knowledge (e.g., $(Obama, Win, Nobel Peace Prize, 2019/10/09)$ can
be mapped as $(PERSON, Win, PRIZE)$).

\subsubsection{Rule Graph.}

A rule graph is a directed graph $\mathrm{G} = \{ \mathrm{V},
\mathrm{E}\}$, where each $v \in \mathrm{V}$ is a node indicating an
atomic rule, and each $e \in \mathrm{E}$ is a rule edge preserving
the sequential relevance between atomic rules. There are two kinds
of rule edges in $\mathrm{E}$. One is derived from the chain
occurring (e.g., $(PERSON, Nominated, PRIZE) \rightarrow (PERSON,
Win, PRIZE)$), termed as $(v_h \rightarrow v_t)$ where $v_h$ is the
head atomic rule and $v_t$ is the tail atomic rule. The other is
derived from the triadic occurring (e.g., $(PERSON, Write, BOOK),
(BOOK$, $Nominated, PRIZE) \rightarrow (PERSON, Win, PRIZE)$),
termed as $((v_h,$ $v_m) \rightarrow v_t)$, where $v_m$ is the
middle atomic rule. By associating atomic rules with rule edges,
paths between atomic rules can describe the occurrence relevance
between two kinds of interactions.

\subsection{Problem Definition}

Detecting anomalies for TKGs that have been offline preserved in the
database is meaningful. However, it is a more valuable but difficult
problem to detect anomalies for TKGs that are online updating,
requiring the model to be efficient, adaptive to online changes, and
easy to rebuild. We term it inductive anomaly detection in TKGs.
\begin{definition}[Inductive anomaly detection in TKGs]
Given an online updating TKG $\mathcal{G}$ where the most recently
updated timestamp is $t_e$, inductive anomaly detection aims to
construct a model $M$ based on $\mathcal{G}$ to find anomalies in
the future timestamps $t > t_e$. It contains classifying whether
each newly arrived knowledge $(s, r, o, t)$ is an anomaly (i.e.,
conceptual or time errors), and determining whether some knowledge
is missing in $t$ (i.e., missing error).
\end{definition}

We leverage the widely used compression principle MDL to construct
$M$, which follows the idea of compression in information theory to
find general patterns to describe the valid data, and thus in turn
reveal anomalies. The sub-problem is hence defined as:
\begin{definition}[Inductive TKG summarization with MDL]
Given an online updating TKG $\mathcal{G}$, we seek to find the
model $M^*$ (i.e., the optimal rule graph) that minimizes the
description length of $\mathcal{G}$,
\begin{equation}
\label{eq:MDL}
    M^* = \mathop{\arg\min}_{M \in \mathcal{M}} \{L(M) + L(\mathcal{G} | M)\}.
\end{equation}
With the constant enrichment of $\mathcal{G}$, $M^*$ varies across
timestamps. However, it is time-consuming to construct $M^*$ from
scratch in every timestamp, requiring a strategy to update $M^*$
incrementally.
\end{definition}

\section{Method}

\subsection{Overview}

\paragraph{\textbf{Motivation.}}

Reflecting on the challenges in TKG anomaly detection, we recognize
that a rule-based summarization approach could effectively tackle
these issues. First, rules encapsulate the most common patterns
within a graph in a human-readable form. If we can map new knowledge
as a set of rules, then they can provide interpretable evidence for
its validity. Second, the complex patterns observed in TKGs stem
from the composition of simpler, independent patterns. If we can
appropriately link these simple rules, then the complex patterns can
be flexibly deduced based on the individual rules. Last, rules
describe the properties of a TKG in a more compact and refined way.
Thus ideally, any semantic and pattern shifts can be described as
modifications of the rules.

\paragraph{\textbf{Solution.}}

\begin{figure}[t]
  \centering
  \includegraphics[width=0.48\linewidth]{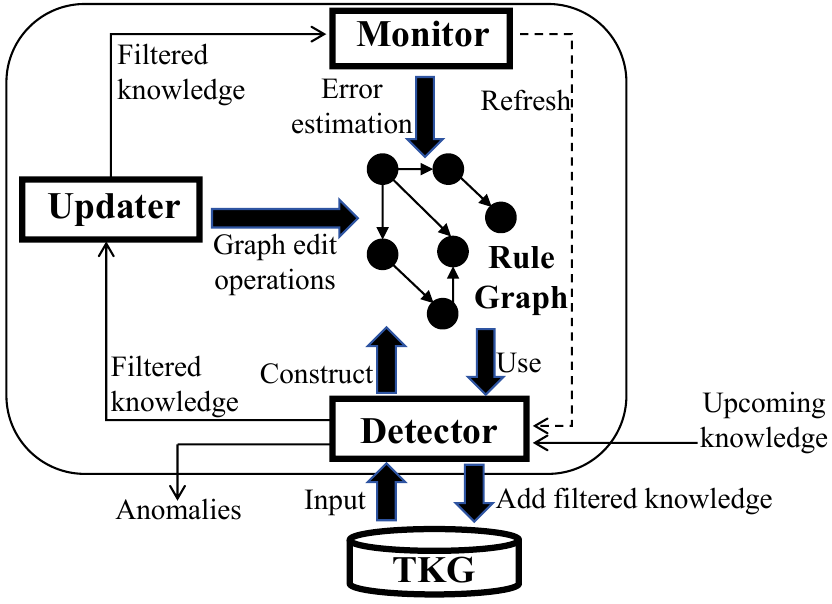}
  \caption{Conceptual illustration of the proposed \textsc{AnoT}
framework.}
  \label{fig:overall}
\end{figure}

In this paper, we propose \textsc{AnoT}, a novel summarization
method for TKG anomaly detection. As depicted in
Figure~\ref{fig:overall}, \textsc{AnoT} takes an online updating TKG
as input, identifies anomalies, and then filters valid knowledge.
The process initiates with the detector module, which constructs a
rule graph based on the offline preserved part of TKG. Upon the
arrival of new knowledge, this module evaluates it against the rule
graph to compute an anomaly score. Subsequently, the updater module
receives valid knowledge identified by the detector module, and then
reforms them as edit operations on the rule graph to handle online
semantic and pattern changes. The monitor module estimates the
approximate error of the rule graph in representing the TKG. When
the approximate error exceeds the threshold, the monitor will inform
the detector to refresh the rule graph based on the current TKG. In
this way, the reachable nodes during walking will give readable
evidence for detection, while the complex patterns can be flexibly
described by the walking paths, and the online changes are uniformly
handled.

In the following, we first define the description length of
$\mathcal{G}$ used to find the optimal rule graph, and then detail
each part of \textsc{AnoT}.

\subsection{Description Length of The Rule Graph}

We employ the minimum description length principle to guide the
construction of the optimal rule graph. In other words, we consider
it as a classic information-theoretic transmitter/receiver setting
\cite{DBLP:conf/nips/WangWPS20}, where the goal is to describe the
graph to the receiver using as few bits as possible. As a result, we
should first define the number of bits required to describe the TKG
(i.e., $L(M)$ and $L(\mathcal{G} | M)$).

\subsubsection{$L(M)$.}

Based on the primitives of MDL principle
\cite{DBLP:journals/datamine/Galbrun22} and the definition of the
rule graph, the encoding cost of a rule graph $\mathrm{G} = \{
\mathrm{V}, \mathrm{E}\}$ consists of the number of atomic rules
$\mathrm{V}$, the number of rule edges $\mathrm{E}$ (both upper
bounded by the number of possible candidates), and the encoding cost
of $V$ and $E$, which is defined as
\begin{equation}
\begin{split}
    L(M) = log(2*|\mathcal{C}_{\mathcal{E}}|^2*|\mathcal{R}|) + log\binom{2*|\mathcal{C}_{\mathcal{E}}|^2*|\mathcal{R}|}{3} + \sum_{v \in \mathrm{V}}L(v)+ \sum_{e \in \mathrm{E}}L(e),
\end{split}
\end{equation}
where the first term is the upper bound of the number of candidate
atomic rules. $|\mathcal{C}_{\mathcal{E}}|$ is the total number of
entity categories derived from function $\mathcal{C}(\cdot)$ (see
Section~\ref{sec:Construct category function}). $|\mathcal{R}|$ is
the number of relations. Each atomic rule has the form of (CATEGORY,
relation, CATEGORY), and thus results in
$|\mathcal{C}_{\mathcal{E}}|^2*|\mathcal{R}|$. Twice because each
relation has two directions. The second term is the upper bound of
the number of candidate rule edges, where $log\binom{A}{B}$ means
the description length of uniformly choosing $B$ elements from $A$
elements. Each rule edge associates two or three atomic rules (i.e.,
chain or triadic occurring), and thus results in $B$=3 as the upper
bound. $L(v)$ and $L(e)$ are respectively the encoding costs of each
atomic rule and each rule edge, defined as
\begin{equation}
\begin{split}
    L(v) = log|\mathcal{C}_{\mathcal{E}}| + (- log \frac{n^{c_s}}{|\mathcal{E}|}) + (- log \frac{n^{c_o}}{|\mathcal{E}|}) + (- log \frac{n^{r}}{|\mathcal{F}|} + 1),
\end{split}
\end{equation}
where the first term is the number of the categories of entities.
The second to fourth terms are the number of bits used to encode
subject categories, object categories, and relations respectively.
Note that we use optimal prefix code
\cite{DBLP:journals/ire/Huffman52} to encode actual categories, so
$n^{c_s}$ is the number of times category $c_s \in
\mathcal{C}_{\mathcal{E}}$ occurs in $\mathcal{G}$ ($n^{c_o}$ is
similar), while $n^{r}$ is the number of times relation $r \in
\mathcal{R}$ occurs in $\mathcal{G}$ and $1$ represents the
direction of the relation. Since the number of relations is constant
across different models, we ignore it during model selection. We
define the encoding cost of each rule edge as
\begin{equation}
\begin{split}
    L(e) = log|\mathrm{E}| + (- log \frac{n^{v_h}}{|\mathrm{E}|}) + (- log \frac{n^{v_m}}{|\mathrm{E}|}) + (- log \frac{n^{v_t}}{|\mathrm{E}|}) + 1,
\end{split}
\end{equation}
where $|\mathrm{E}|$ is the number of the rule edges and $n^{v_h}$
is the number of head atomic rule $v_h$ occurs in rule graph
$\mathrm{G}$. $1$ represents the direction of the edge. Note that
for brevity, we only give $L(e)$ of the triadic occurring and the
chain occurring can be easily extended by removing the auxiliary
atomic rule part (i.e., the third term).

\subsubsection{$L(\mathcal{G} | M)$.}
\label{sec:L(g|M)}

Each atomic rule can describe a set of facts in TKG (e.g., atomic
rule $(PERSON, Wins,$ $PRIZE)$ can describe fact $(Obama, Win, Nobel
Peace Prize, 2019/10/09)$), and each rule edge can describe a set of
occurring relationships among facts. For example, rule edge
$(PERSON, Win the Selection,$ $COUNTRY) \rightarrow (PERSON,
President of, COUNTRY)$ can describe the relationship that fact
$(Obama,$ $President of, United States, 2009/01/20)$ occurs
subsequently after the fact $(Obama, Win the Selection,$ $United
States, 2008/11/04)$. These described facts are called
\textbf{correct assertions}. They can be encoded by the given rule
graph. Moreover, TKGs inevitably contain noise and uncommon facts
and thus there may be facts that cannot be encoded by the rule
graph, called \textbf{negative errors}. Therefore, the encoding cost
of $\mathcal{G}$ by the rule graph $M$ is $L(\mathcal{G} | M) =
L(\mathcal{A}^{\mathcal{G}}) + L(\mathcal{N}^{\mathcal{G}})$.
$L(\mathcal{A}^{\mathcal{G}}) $ is the encoding cost of the correct
assertions and $L(\mathcal{N}^{\mathcal{G}})$ is the encoding cost
of the negative errors. The encoding cost of the correct assertions
is defined as
\begin{equation}
\begin{split}
    L(\mathcal{A}^{\mathcal{G}}) = \sum_{v \in V} \sum_{a_v \in \mathcal{A}_v^{\mathcal{G}}} L(a_v) + \sum_{e \in E} \sum_{a_e \in \mathcal{A}_e^{\mathcal{G}}} L(a_e),
\end{split}
\end{equation}
where $\mathcal{A}_v^{\mathcal{G}}$ and
$\mathcal{A}_e^{\mathcal{G}}$ are the sets of correct assertions of
each atomic rule $v$ and each rule edge $e$ respectively. $L(a_v)$
and $L(a_e)$ are respectively the encoding costs of fact $a_v$
(i.e., a correct assertion of $v$) and relationship $a_e$ (i.e., a
correct assertion of $e$), defined as
\begin{equation}
\begin{split}
\label{eq:l(a_v)}
    L(a_v) = (-log\frac{n^{s_v}}{|\mathcal{A}_v^{\mathcal{G}}|}) + (-log\frac{n^{o_v}}{|\mathcal{A}_v^{\mathcal{G}}|}),
\end{split}
\end{equation}
\begin{equation}
\begin{split}
    L(a_e) = (-log\frac{n^{a^h_e}}{|\mathcal{A}_e^{\mathcal{G}}|}) + (-log\frac{n^{a^m_e}}{|\mathcal{A}_e^{\mathcal{G}}|}) + (-log\frac{n^{a^t_e}}{|\mathcal{A}_e^{\mathcal{G}}|}),
\end{split}
\end{equation}
where $n^{s_v}$ and $n^{o_v}$ are respectively the numbers of times
entity $s$ and $o$ occur in all the correct assertions of atomic
rule $v$ (i.e., $\mathcal{A}_v^{\mathcal{G}}$). Note that the
encoding cost of the relation is ignored since it has been included
in $L(v)$. $a^h_e$, $a^m_e$, and $a^t_e$ are respectively the head
fact, middle fact, and the tail fact of $a_e$, and thus $n^{a^h_e}$
is the number of times fact $a^h_e$ occurs in all the correct
assertions of rule edge $e$ (i.e., $\mathcal{A}_e^{\mathcal{G}}$).
$a^m_e$ and $a^t_e$ are defined similarly.

The negative errors contain two parts: the facts that cannot be
mapped into any atomic rules, and the facts that can be mapped but
cannot be associated with any other facts. Given the adjacency
tensor $A_t$ of each timestamp, the unmodeled facts can be defined
as $A_t^- = A_t - A_t^m$. $A_t^m$ is a subset of $A_t$ where each
element is set as 1 if 1) it is 1 in $A_t$; 2) the corresponding
fact can be mapped into atomic rules; and 3) the corresponding fact
can be associated with a previous fact in $t' < t$ via the rule
edges. Since the number of missing facts can be inferred given the
total number of facts and the number of facts that are already
explained by the model, we only consider the cost of encoding the
positions of 1 in $A_t^-$. The encoding cost of the negative errors
can be then defined as
\begin{equation}
\begin{split}
\label{eq:L(N^g)}
    L(\mathcal{N}^{\mathcal{G}}) = \sum_{t \in \mathcal{T}} log\binom{|{\mathcal{E}}|^2*|\mathcal{R}| - |A_t^m|}{|A_t^-|},
\end{split}
\end{equation}
where $|\cdot|$ is set cardinality and the number of 1s in a tensor.
As the above definitions give us a reliable way to measure the
quality of summarizing a TKG, we then aim to find the best TKG
summarization model by minimizing such description length.

\begin{figure}[t]
  \centering
  \includegraphics[width=0.68\linewidth]{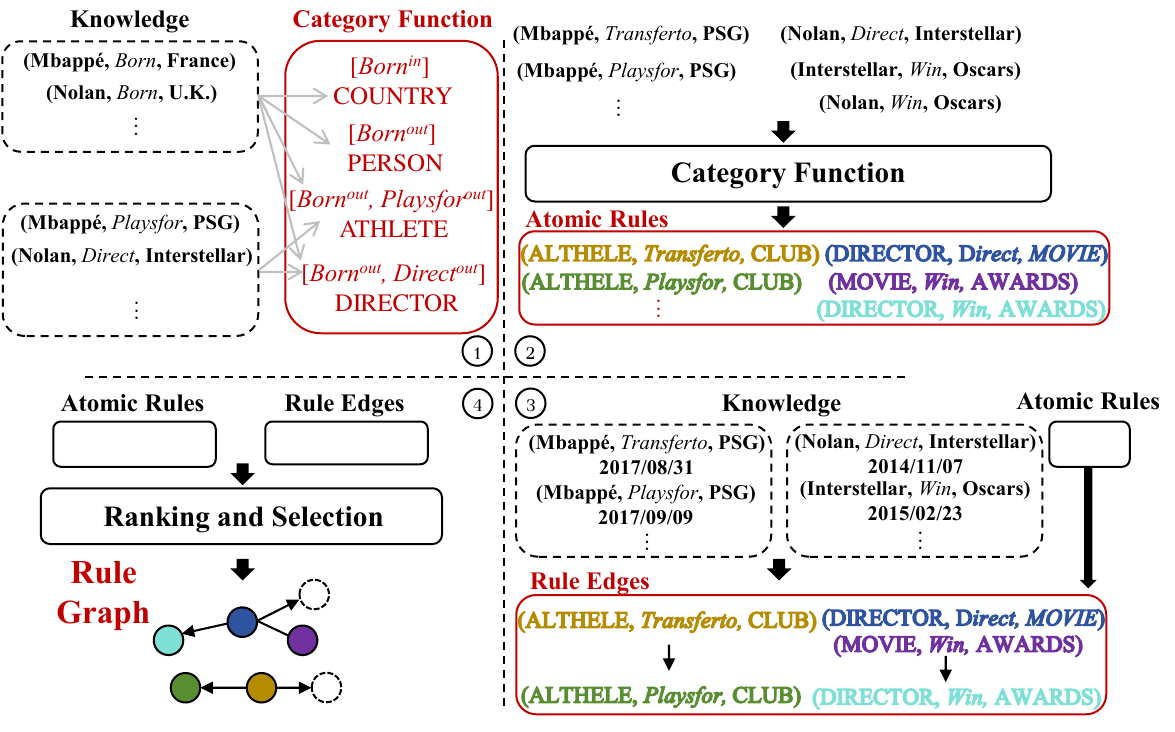}
  \caption{Rule graph construction, which contains: 1. Category
function construction, 2. Generating candidate atomic rules, 3.
Generating candidate rule edges, 4. Ranking and selecting.}
  \label{fig:construct}
\end{figure}

\subsection{Detector}

\begin{figure}[t]
  \centering
  \includegraphics[width=0.68\linewidth]{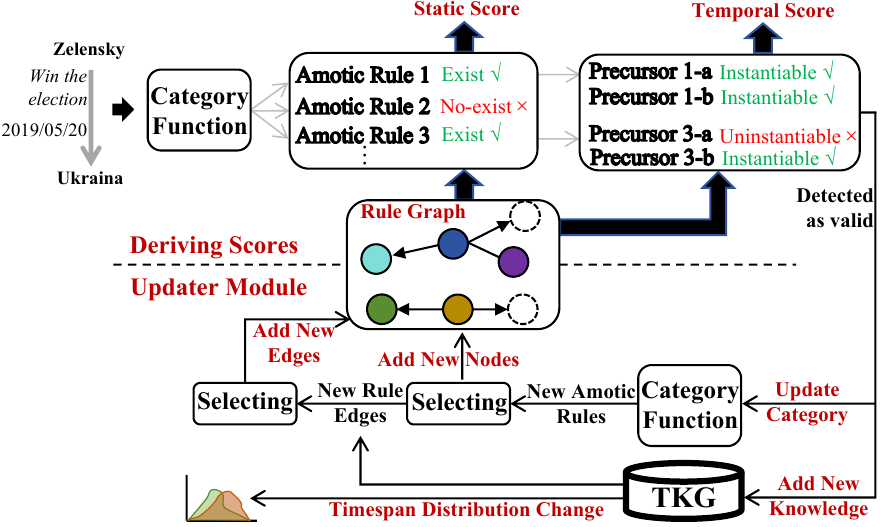}
  \caption{Conceptual illustrations of the scoring process and
the updater module.}
  \label{fig:scoring_updater}
\end{figure}

The detector module has two functions: \textbf{1) Construct the
optimal rule graph.} As shown in Figure~\ref{fig:construct}, a
category function is first constructed based on existing knowledge
(Section~\ref{sec:Construct category function}), and then it will be
used to map knowledge as candidate atomic rules and candidate rule
edges (Section~\ref{sec:Candidate Generation}). Finally, the most
expressive candidates will be iteratively selected to construct the
rule graph (Section~\ref{sec:ranking and selecting}). More details
can be found in Algorithm~\ref{alg:constrcut}. \textbf{2) Generate
anomaly scores.} As shown in the upper part of
Figure~\ref{fig:scoring_updater}, new knowledge will be first mapped
as a set of atomic rules. Atomic rules that exist in the rule graph
can give evidence of their conceptual validity, which will derive
the static scores (Section~\ref{sec:scoring}). Then, the evidence of
time validity is gathered by recursively walking on the rule graph
and instantiating the processor nodes, which will derive the
temporal scores. More details can be found in
Algorithm~\ref{alg:score}.

\begin{algorithm}[t] \footnotesize
    \caption{Construct the rule graph}
    \begin{algorithmic}[1]
        \State \textbf{Input:} Offline preserved part of TKG $\mathcal{G}$
        \State \textbf{Output:} A model $M^*$ consisting of atomic rules and rule edges
        \State Read $\mathcal{G}$ and construct the category function \Comment{\ref{sec:Construct category function}}
        \State Generate candidate atomic rules $\mathcal{P}(v)$ and candidate rule edges $\mathcal{P}(e)$ based on $\mathcal{G}$ and the category function \Comment{\ref{sec:Candidate Generation}}
        \State Rank all $v \in \mathcal{P}(v)$ first by $\Delta L(\mathcal{G} | M \cup \{v\}) \downarrow$ then by $|\mathcal{A}_v^{\mathcal{G}}| \downarrow$ and finally by $ID \downarrow$ \Comment{\ref{sec:ranking and selecting}}
        \State Rank all $e \in \mathcal{P}(e)$ first by $\Delta L(\mathcal{G} | M \cup \{e\}) \downarrow$ then by $|\mathcal{A}_e^{\mathcal{G}}| \downarrow$ and finally by $ID \downarrow$ \Comment{\ref{sec:ranking and selecting}}
        \State $M^* \leftarrow \emptyset$
        \While{not converged} \Comment{\ref{sec:ranking and selecting}}
        \For{$v \in \mathcal{P}(v)$}
        \If{$L(\mathcal{G} | M^* \cup {v}) < L(\mathcal{G} | M^*)$ }
        \State $M^* \leftarrow M^* \cup {v}$ \Comment{add more $v$ to $M^*$}
        \EndIf
        \EndFor
        \For{$e \in \mathcal{P}(e)$}
        \If {$L(\mathcal{G} | M^* \cup {e}) < L(\mathcal{G} | M^*)$ }
        \State $M^* \leftarrow M^* \cup {e}$ \Comment{add more $e$ to $M^*$}
        \EndIf
        \EndFor
        \EndWhile
    \end{algorithmic}
    \label{alg:constrcut}
\end{algorithm}

\subsubsection{Construct Category Function.}
\label{sec:Construct category function}

Entity categories are often missing in real-world TKGs
\cite{DBLP:conf/nips/YouMDKL20}. Fortunately, we find that the
category of an entity is largely related to the relations it
interacts with. For example, an entity that interacts with relations
$BornIn$ and $PlaysFor$ should have a category of $ATHLETE$, and an
entity that interacts with relations $BornIn$ and $Create$ can be an
$ARTIST$. Intuitively, the more relations are considered, the more
fine-grained information a category can imply. This inspires us to
generate categories for entities by extracting the frequent relation
combinations. A relation combination occurs more frequently across
entities means it can describe a more general property shared by
entities, which is more likely to imply a category.

Formally, we define the interaction relation set of an entity $e \in
\mathcal{E}$ as $R(e) = \{ r| (e, r, o, t) \in \mathcal{F} \}$. With
each entity providing a relation set, we then use the PrefixSpan
algorithm \cite{DBLP:conf/icde/PeiHPCDH01} to find the most frequent
subset within $R(e)$ across all entities. Each identified subset is
a frequent relation combination that suggests a potential category.
Subsets encompassing more relations imply more granular categories.
However, finding frequent subsets with large sizes is very
time-consuming. To counterbalance efficiency with categories'
quality, we propose to only find small relation combinations (i.e.,
up to 3 relations) and then iteratively aggregate selected subsets.
Let the output of PrefixSpan be $\textbf{R} = \{ (R_1, E_1), (R_2,
E_2),... \}$, where $R_i$ is the ith most frequent relation
combination $(r_i^1, r_i^2, r_i^3)$ and $E_i = \{ e| e\in
\mathcal{E}, R_i \subseteq R(e) \}$ is the entities encompassed by
$R_i$. We commence with \textbf{entity-based aggregation}: if a
significant overlap exists between entities in $E_i$ and $E_j$
(exceeding 90\%), it indicates shared properties that necessitate
simultaneous description via both relation combinations.
Consequently, we introduce $(R = R_i \cup R_j, E = E_i \cap E_j)$
into $\textbf{R}$ as a more fine-grained category. We then perform
\textbf{relation-based aggregation}: if a significant overlap exists
between relations in $R_m$ and $R_n$ (exceeding 90\%), it means the
categories implied by $R_m$ and $R_n$ are very similar. Therefore,
we add $(R = R_m \cap R_n, E = E_m \cup E_n)$ into $\textbf{R}$ as a
more generalizable category. These aggregation steps are circularly
executed until no further combinations can be aggregated. Following
this phase, each relation combination $R_i$ is conceptualized as an
implicit category $c_i$, with the respective entities in $E_i$ being
categorized accordingly. To reduce category redundancy, we sort the
relation combinations in descending order of the number of their
covered entities and select one by one until each entity has at
least $k$ categories.

\subsubsection{Candidate Generation.}
\label{sec:Candidate Generation}

To construct the rule graph, we should first generate all possible
rules and rule edges as candidates based on the input TKG. For each
fact $(s, r, o, t)$, we generate the corresponding candidate atomic
rules as $\{(c_i, r, c_j) | c_i \in C(s), c_j \in C(o)\}$, where
$C(s)$ is the category set of $s$. We gather the rules derived from
all facts $(s, r, o, t) \in \mathcal{F}$ as the candidate set of
atomic rules.

To generate all possible chain-occurring-based rule edges, we first
construct the interaction sequence $S(s,o) = \{ r_1, r_2, ...\}$ for
each entity pair appeared in $\mathcal{G}$. The interaction sequence
preserves all the relations that occurred between $s$ and $o$ and is
sorted by the ascending order of their occurrence timestamps. Thus,
any adjacent relations in the sequence represent two interactions
between $s$ and $o$ that occur successively, which may imply a
chain-occurring pattern. Formally, given each entity pair $(s,o)$
and its corresponding interaction sequence, we generate the
candidate rule edges as $\{ (c_s, r_m, c_o) \rightarrow (c_s, r_n,
c_o) | c_s \in C(s), c_o \in C(o), r_m, r_n \in S(s, o), m < n \}$.
We gather the rule edges derived from all entity pairs that appear
in $\mathcal{G}$ as the candidate set of chain-occurring-based rule
edges. Since the occurrence timespan between different relations may
vary, e.g., $Make Statement$ may occur a few days after $Win the
Election$, but $Retirement$ may occur several years later,
considering the timespan of occurrence between two relations assists
in determining the occurrence of a fact at a particular timestamp.
Therefore, we also preserve the occurrence timespans of facts for
each rule edge $e$ (e.g., $t_n - t_m$ for chain-occurring facts $(s,
r_m, o, t_m)$ and $(s, r_n, o, t_n)$) and results in a timespan set
$T(e)$.

For triadic-occurring-based rule edges, in each timestamp $t \in
\mathcal{T}$, we find facts that occur at $t$ and share one same
entity (e.g., $(s, r_m, o, t)$ and $(h, r_n, o, t)$). Then, we find
the most closely occurred fact in $t_p \geq t$ that contains $s$ and
$h$ (e.g., $(s, r_p, h, t+3)$). These three facts describe the
formation process of a triadic closure, which may imply a
triadic-occurring pattern. Since there is local randomness in the
occurrence time of facts \cite{DBLP:conf/sigmod/MondalD12}, we relax
the same time restriction of the former two facts as co-occurring
within a short period. Formally, for each timestamp $t$, the
triadic-occurring-based candidate rule edges are generated as $\{
((c_s, r_m, c_o), (c_h, r_n, c_o)) \rightarrow (c_s, r_p, c_h) | c_s
\in C(s), c_o \in C(o), c_h \in C(h), |\mathcal{T}(e_s, r_m, e_o) -
t| \leq L, |\mathcal{T}(e_h, r_n, e_o) - t| \leq L, \mathcal{T}(e_s,
r_p, e_h) \geq t+L \}$, where $\mathcal{T}(\cdot)$ is the occurrence
timestamp of fact and $L$ is a hyperparameter. We also preserve the
occurrence timespans for these facts.

\subsubsection{Ranking and Selection.}
\label{sec:ranking and selecting}

We propose a greedy approach to select the most representative
candidate into the rule graph iteratively. Our objective is to
select the candidate that leads to the largest encoding cost
reduction in each iteration. Recognizing that varying orders of
selection may yield inconsistent models, thus affecting
reproducibility, we implement a structured ranking mechanism that
ensures a consistent selection order of candidates. Since the more a
candidate can reduce negative errors, the more valuable it might be,
we first rank candidates based on the descending order of their
error reduction $\Delta L(\mathcal{G} | M \cup \{x\}) =
L(\mathcal{G} | M ) - L(\mathcal{G} | M \cup \{x\})$, where $x$
represents candidate atomic rule $v$ or rule edge $e$. The ties in
the ranking are broken by selecting candidates with more correct
assertions. The final tie-breaker is the ID of each candidate. We
separately rank atomic rules and rule edges since they have
different magnitudes in the cost reduction. The ranked candidate
rules and rule edges are respectively termed as $\mathcal{P}(v)$ and
$\mathcal{P}(e)$.

After ranking the candidates, $M$ is initialized as $\emptyset$ and
each $ v \in \mathcal{P}(v)$ is first selected in ranked order. For
each $v$, we compute the description length when $v$ is added into
$M$ (i.e., $L(\mathcal{G}, M \cup \{v\})$). If it is less than
$L(\mathcal{G}, M)$, $v$ can enhance the expressive capability of
$M$. Thus, we add $v$ into $M$. We perform the selection passes over
$\mathcal{P}(v)$ until no new atomic rules can be added. We then
perform the same selection process on $\mathcal{P}(e)$ to add rule
edges into $M$. Note that some selected rule edges may contain
atomic rules that are not selected in the former process. We
restrict the usage of these atomic rules only to verify the time
errors. The obtained approximately optimized rule graph is termed as
$M^* = \{\mathrm{V}^*, \mathrm{E}^* \}$.

\subsubsection{Deriving Anomaly Scores.}
\label{sec:scoring}

\begin{algorithm}[t] \footnotesize
    \caption{Derive the anomaly scores}
    \begin{algorithmic}[1]
        \State \textbf{Input:} New knowledge $(s, r, o, t)$ and rule graph $M^*$
        \State \textbf{Output:} Static score $\mathbb{S}(s, r, o, t)$ and temporal score $\mathbb{T}(s, r, o, t)$
        \State Generate mapped atomic rule set $\mathrm{V}(s, r, o, t)$
        \State $temp_s \leftarrow 0$
        \For{$v \in \mathrm{V}(s, r, o, t)$} \Comment{Eq.~\ref{eq:s_score}}
        \State $tmp_s \leftarrow tmp_s + |\mathcal{A}_v^{\mathcal{G}}|$
        \EndFor
        \State $\mathbb{S}(s, r, o, t) = \frac{1}{tmp_s}$
        \If{$tmp_s < \lambda$} \Comment{$\lambda$ is a threshold}
        \State return $\mathbb{S}(s, r, o, t)$
        \EndIf
        \State $V' \leftarrow \mathrm{V}(s, r, o, t), tmp_t \leftarrow 0, MAX\_STEP \leftarrow 0$
        \State $tmpList \leftarrow \emptyset$
        \For{$v \in V'$} \Comment{Eq.~\ref{eq:t_score}}
        \For{$v_i \in \mathcal{N}_{in}(v)$}
        \If{$v_i$ is instantiable}
        \State $tmp_t \leftarrow tmp_t + \frac{|\mathcal{A}_v^{\mathcal{G}}|}{\theta + 1}$
        \Else
        \State $tmpList \leftarrow tmpList \cup {v_i}$
        \EndIf
        \EndFor
        \EndFor
        \If{$MAX\_STEP < K$} \Comment{$K$ is a hyper-parameter}
        \State $MAX\_STEP \leftarrow MAX\_STEP + 1$
        \State $V' \leftarrow tmpList$
        \State Go to 11
        \EndIf
        \State $\mathbb{T}(s, r, o, t) = \frac{1}{tmp_t}$
    \end{algorithmic}
    \label{alg:score}
\end{algorithm}

Intuitively, nodes and edges in the rule graph explain the common
patterns of knowledge occurring in TKG. Thus, new knowledge that
cannot be mapped as nodes or cannot be associated with other
knowledge via edges is unexplained and likely to be anomalous. We
make this intuition more principled by defining static scores and
temporal scores for tuples.

\textbf{Static scores.} Nodes in $M^*$ represent valid interaction
patterns found in the TKG. When new knowledge is mapped to a node in
$M^*$, it means that the knowledge can be explained by an observed
pattern, supporting its conceptual validity. The validity of new
knowledge is proportionate to the number of nodes it mapped to.
Furthermore, some atomic rules can explain more knowledge and thus
give stronger evidence. We define the static scores as
\begin{equation}
\begin{split}
\label{eq:s_score}
    \mathbb{S}(s, r, o, t) = \frac{1}{\sum_{v \in \mathrm{V}^*(s, r, o, t)} |\mathcal{A}_v^{\mathcal{G}}|},
\end{split}
\end{equation}
where $\mathrm{V}^*(s, r, o, t)$ is the set of nodes that new
knowledge $(s, r, o, t)$ can be mapped to.
$|\mathcal{A}_v^{\mathcal{G}}|$ is the number of correct assertions
of $v$. The higher $\mathbb{S}$ means it is more likely to be a
conceptual error.

\textbf{Temporal scores.} Each in-coming edge of a node $v$ in the
rule graph provides an inducement for the interaction represented by
$v$ to occur, while the timespans preserved in the edge provide the
prompt of when it should occur. Therefore, we propose to walk on the
rule graph starting from the mapped nodes of new knowledge to find
evidence for it to occur. Specifically, given the new knowledge $(s,
r, o, t)$, we first map it to nodes in the rule graph by
conceptualizing it as a set of atomic rules $\mathrm{V}^*(s, r, o,
t)$. Then, for each $v \in \mathrm{V}^*(s, r, o, t)$, we find all of
its in-coming edges (i.e., $(v_i \rightarrow v)$ or $((v_i, v_k)
\rightarrow v)$) and gather all precursor nodes (i.e., $\{ v_i |
(v_i \rightarrow v) \in \mathrm{E}^* or ((v_i, v_k) \rightarrow v)
\in \mathrm{E}^* \}$). We then perform the instantiate on these
precursor nodes. For each $v_i$, it is instantiable if there is a
fact in $v_i$'s correct assertions that can form an occurring
relationship (i.e., chain or triadic occurring) with new knowledge
$(s, r, o, t)$. The instantiable precursor nodes give evidence for
the occurrence of new knowledge. Note that TKG inevitably contains
noise such as knowledge missing, which can cause node instantiation
to fail. We propose a recursive strategy to enhance the robustness
of our scoring. If precursor node $v_i$ fails to be instantiated, we
find all the precursor nodes of $v_i$ and use the instantiation of
them as alternative evidence from $v_i$. Our algorithm will traverse
the rule graph depth-first to gather evidence until the maximum
number of hops is reached. The temporal score is formally defined as
\begin{equation}
\begin{split}
\label{eq:t_score}
    \mathbb{T}(s, r, o, t) = \frac{1}{\sum_{v \in \mathrm{V}^*(s, r, o, t)} \sum_{v_i \in \mathcal{N}_{in}(v)} x},
\end{split}
\end{equation}
where $x = \frac{|\mathcal{A}_v^{\mathcal{G}}|}{\theta + 1}$ if
$v_i$ is instantiable, else $x = \sum_{v_j \in
\mathcal{N}_{in}(v_i)} x$ where $\mathcal{N}_{in}(v)$ is the set of
in-coming neighbors of $v$. $\theta = |\{ \tau_j | \tau_j \in T((v_i
\rightarrow v)), | \tau_j - |t - t_i|| \leq L\}|$ which indicates
the gap between the timespan of the instantiations and the preserved
timespans. $t_i$ is the occurrence timestamp of the instantiated
previous knowledge. Temporal scores can be further extended by
adding the number of instantiable out-coming edges to the numerator
of Eq.~\ref{eq:t_score}. Specifically, each out-coming edge of node
$v$ describes an interaction that should occur after $v$. Therefore,
if it can be instantiated by previous knowledge, it means the
occurrence of new knowledge violates a common occurrence order. The
higher $\mathbb{T}$ means it is the more likely to be a time error.
Meanwhile, if knowledge gets both low $\mathbb{S}$ and $\mathbb{T}$
but is not preserved in TKG, it is likely to be a missing error.

\textbf{Correcting prompts.} For conceptual errors, we use the
atomic rules that can partially describe the anomaly knowledge $(s,
r, o, t)$ as its correcting prompts, e.g., $\{ (c_s, r, c_e) | c_e
\in \mathcal{C}(e), e \in \mathcal{E}, (c_s, r, c_e) \in V^* \}$ and
$\{ (c_s, r_i, c_o) | c_s \in \mathcal{C}(s), c_o \in
\mathcal{C}(o), (c_s, r_i, c_o) \in V^* \}$, which tell us how to
revise the entity or relation in anomaly knowledge to make it valid.
For time errors, the instantiable in-coming edges (evidence of
correctness) and instantiable out-coming edges (evidence of anomaly)
give prompts of when the new knowledge occurs is appropriate (i.e.,
maximize the instantiable in-coming edges and minimize the
instantiable out-coming edges). During walking, precursor nodes that
fail to instantiate may indicate a missing knowledge, and thus give
us prompts to extract new knowledge.

\textbf{Case demonstration.} Here we demonstrate how our strategies
detect anomalies in Figure~\ref{fig:anomalies}. $(Kim Jong UN, Held
Military Exercises, South Korea, 2020/08/23)$ will be assigned with
a high $\mathbb{S}$ due to the interaction preference conflict of
category $PRESIDENT$, and thus be detected a conceptual error.
$(Trump, President of, United States, 2023/01/20)$ will be assigned
with a high $\mathbb{T}$ since it has occurrence order conflict with
$(Trump, Outgoing President, United States, 2021/01/19)$, and thus
be a time error. During temporal scoring, \textsc{ANoT} needs to
traverse the rule graph to find instantiateable nodes, and $(PERSON,
Win the Election, United States)$ will be found uninstantiateable.
By verifying the low $\mathbb{S}$ and $\mathbb{T}$ of instantiating
it using $Joe Biden$, this knowledge will be detected as a missing
error.

\subsection{Updater}

\begin{algorithm}[t] \footnotesize
    \caption{Update the rule graph}
    \begin{algorithmic}[1]
        \State \textbf{Input:} New valid knowledge $(s, r, o, t)$ and rule graph $M^*$
        \State \textbf{Output:} The updated rule graph $M^*$
        \State $\mathcal{G} \leftarrow \mathcal{G} \cup \{(s, r, o, t)\}$ \Comment{handle graph structure changes}
        \For{$e \in (s, o)$}
        \If{$r \notin R(e)$} \Comment{handle entity semantic changes}
        \State $R(e) \leftarrow R(e) \cup \{r\}$
        \State Generate candidate category set $R_c(e) = \{R_i | r \in R_i, R(e) \cap R_i \neq \emptyset\}$
        \State Find $c_i \in R_c(e)$ with the maximum $E_i$
        \State $\mathcal{C}(e) \leftarrow \mathcal{C}(e) \cup \{c_i\}$
        \EndIf
        \EndFor
        \For{$v \in \{ (c_s, r, c_o) | c_s \in \mathcal{C}(s), c_o \in \mathcal{C}(o), (c_s, r, c_o) \notin V^* \}$} \Comment{handle new interaction patterns}
        \If{$L(\mathcal{G} | M^* \cup {v}) < L(\mathcal{G} | M^*)$}
        \State $M^* \leftarrow M^* \cup \{v\}$
        \For{$f' \in \{ (s, r', o, t') | (s, r', o, t') \in \mathcal{G}, |t-t'| \leq L\}$}
        \State $M^* \leftarrow M^* \cup \{ v \rightarrow v' \}$ \Comment{$v'$ is the atomic rule that can describe $f'$}
        \State $T(e) \leftarrow T(e) \cup \{ t - t'\}$ \Comment{handle timespan distribution changes}
        \EndFor
        \EndIf
        \EndFor
    \end{algorithmic}
    \label{alg:update}
\end{algorithm}

The continuous enrichment of TKG will alter the graph structure,
entity semantics, and graph patterns, and introduce new entities,
requiring \textsc{ANoT} to adapt online. As shown in the bottom part
of Figure~\ref{fig:scoring_updater}, we propose an updater module to
flexibly handle these changes. Based on Algorithm~\ref{alg:update},
we detail the updater as follows.

\textbf{Graph structure changes.} $(s,r,o,t)$ will first be added
into $\mathcal{G}$. Thus, the next time the detector module needs to
instantiate rules and rule edges, it can access the latest version
of $\mathcal{G}$, and thus adapts the scoring to graph structure
changes brought by new knowledge.

\textbf{Entity semantic changes.} New knowledge can change an
entity's semantics if it includes a relation the entity has not
interacted with before (e.g., relation $President of$ will add a new
category $PRESIDENT$ for a person). This inspires us to handle
entity semantic changes via category editing. When new relation $r$
is introduced for entity $s$, relation combinations $\{R_i | r \in
R_i, R(s) \cap R_i \neq \emptyset\}$, which may describe the new
categories of $s$, will be identified first. Then, the relation
combination with the most covered entities will be added to the
category function as a new anonymous category for $s$.

\textbf{Graph pattern changes.} New categories may derive atomic
rules that do not exist in the current rule graph, indicating an
emerging interaction pattern. We model such pattern changes as the
enrichment on the node set of $M^*$. For each $v \in \{ (c_s, r,
c_o) | c_s \in \mathcal{C}(s), c_o \in \mathcal{C}(o), (c_s, r, c_o)
\notin V^* \}$ derived from new knowledge, we calculate the encoding
cost $L(\mathcal{G}, M^* \cup \{v\})$ for it and if $L(\mathcal{G},
M^* \cup \{v\})$ is less than $L(\mathcal{G}, M^*)$, $v$ will be
added into $\mathrm{V}^*$. We then build new edges for the new node
by finding previous knowledge in the form of $(s, r_i, o, t_j)$
where $t_j \leq t$. Considering the time consumption, we only
perform chain-based associations. The other graph pattern change
appears in the occurrence timespan distribution. For example, a
regular consultation mechanism established between two countries
will change the timespans of relation $Consult$ that appears between
these two entities. Therefore, for each rule edge $e$ that can
describe the new knowledge, we also add the timespans of its newly
described facts into $T(e)$, and thus the calculation of the
temporal score can adapt to the changes of occurring timespans.

\textbf{New entities.} For new entities brought by new emerging
knowledge, we follow the same process as entity semantic changes to
generate new categories for them, and thus knowledge that contains
new entities can also be mapped into our rule graph and be handled
uniformly. With the continuous emergence of new knowledge deriving
the update of the rule graph, \textsc{ANoT} can keep on learning new
patterns and thus adapt to new emerging knowledge.

Although new knowledge may have conflicts with some existing rules
or do not fit them, it can still be considered valid since our
counting-based scoring does not require all rules to be met. These
conflicts and mismatches may indicate patterns not observed before.
Our updater module aims to filter and keep these patterns in the
rule graph, so they can help explain future knowledge.

\subsection{Monitor}

The error of the rule graph may accumulate with the updating. Thus,
it is necessary to refresh the model at an appropriate time. There
could be some heuristic strategies, such as restarting after a
certain time or restarting after a certain number of new knowledge.
However, they do not always work since the error accumulation does
not change uniformly with time, and there are no shared trends of
error accumulation across different TKGs
\cite{DBLP:conf/semweb/XuNAYL20}. An untimely restart will lead to
excessive error and reduce the detection accuracy and too frequent
restart will lead to low efficiency. Fortunately, the encoding cost
of negative errors defined in Eq.~\ref{eq:L(N^g)} gives us an
information-theoretic metric to measure the availability of the rule
graph. It has three advantages: 1) It is easy to calculate and thus
will not affect the detection efficiency. 2) It is discretely
calculated based on each new knowledge, avoiding the negative
impacts caused by uneven knowledge distribution. 3) It is a
data-driven metric and thus can adapt to different domains. We
define the metric as
\begin{equation}
\begin{split}
\label{eq:L(N^g)}
    L(\mathcal{N}^{\mathcal{G}^o}) = \sum_{t_i > t_e} log\binom{|{\mathcal{E}}|^2*|\mathcal{R}| - |A_{t_i}^m|}{|A_{t_i}^-|},
\end{split}
\end{equation}
where $t_e$ is the latest timestamp of the offline preserved part of
$\mathcal{G}$. Our monitor module calculates
$L(\mathcal{N}^{\mathcal{G}^o})$ at each new timestamp. If
$L(\mathcal{N}^{\mathcal{G}^o}) > L(\mathcal{N}^{\mathcal{G}})$, it
means that the current model has performed worse on unseen data than
seen data. Thus, the monitor will call the detector to reconstruct
the rule graph based on the current TKG.

\subsection{Complexity Analysis}

Generating candidate atomic rules involves iterating each knowledge
and its entities' categories. The number of candidate atomic rules
generated by each edge $(s, r, o, t)$ is
$|\mathcal{C}(s)|*|\mathcal{C}(o)|$. Letting $\mathcal{C}_{max}$ be
the max number of categories over all entities, the complexity of
generating candidate atomic rules is
$O(\mathcal{C}_{max}^2*|\mathcal{F}|)$. Generating candidate rule
edges needs to iterate knowledge within a timespan. Letting
$f_{max}$ be the max number of previous knowledge that can be
associated, the complexity of generating candidate rule edges is
$O(f_{max}^2*|\mathcal{F}|)$, where $f_{max}^2$ is brought by the
triadic closure searching. Since the complexity of computing the
error cost $L(\mathcal{G}|M)$ is constant, the complexities of
ranking and selecting rules and edges are $O(|\mathcal{P}(v)|
log|\mathcal{P}(v)|) $ and $O(|\mathcal{P}(e)|
log|\mathcal{P}(e)|)$, where $|\mathcal{P}(v)|$ and
$|\mathcal{P}(e)|$ are the numbers of candidate atomic rules and
rule edges. The scoring process and the updater module need to
traverse the category of an entity and find the associated
knowledge, leading to complexities as $O(|C(s)|+|C(o)|+f_{max})$.
The monitor only requires computing
$L(\mathcal{N}^{\mathcal{G}^o})$, which is a small constant.

\subsection{Generalize to Time-duration-based TKGs}

Different from timestamp-based facts discussed before, each fact
$(s, r, o, t_{start}, t_{end})$ in time-duration-based TKGs
\cite{DBLP:conf/emnlp/DasguptaRT18} is associated with start and end
timestamps to indicate its valid duration, e.g., $(Bill Gates,
Married with, Melinda, 1994/1/1, 2021/5/3)$. Similarly, conceptual,
time and missing errors can also exist in such TKGs.

\textbf{Conceptual errors.} Time-duration-based and timestamp-based
facts only differ in their time annotations. Since the detection of
conceptual errors only relies on finding the conflicts of
interaction preference for entities and not related to time
information, the proposed static score can be seamlessly used in
time-duration-based TKGs.

\textbf{Time errors.} Time-duration-based facts can be invalid due
to delays or errors in extracting start and end timestamps, causing
conflicts with the timestamps of other valid facts, e.g., the
relation $Chairman of$ should not start after the $Works at$
relation ends between a person and a company. Therefore, time errors
can be detected by finding such conflicts. Since rule edges indicate
that the tail atomic rule should follow the head atomic rule, we can
create four types of rule graphs for each time-duration-based TKG by
generating rule edges using different time annotation combinations:
\begin{compactitem}
    \item \textbf{ST.-ST.} This rule graph is generated by
only considering $t_{start}$ of facts. Each edge describes that an
atomic rule should start after the other one has started. Facts are
first transferred to timestamp-based by only preserving $t_{start}$
and then Algorithm \ref{alg:constrcut} is used for construction.
    \item \textbf{ED.-ED.} This rule graph is generated by
only considering $t_{end}$ during construction. Each edge describes
that an atomic rule should end after the other one has ended.
    \item \textbf{ST.-ED.} Each edge in this rule graph
indicates that one atomic rule should end after another has started.
During associating atomic rules (Section~\ref{sec:Candidate
Generation}), $t_{end}$ is used for tail fact and $t_{start}$ is
used for head fact.
     \item \textbf{ED.-ST.} Each edge in this rule graph
describes that an atomic rule should start after the other one has
ended. Similarly, $t_{start}$ of a fact will be used if it serves as
a tail fact, and $t_{end}$ will be used if it is a head fact.
\end{compactitem}
Given these rule graphs, the scoring process
(Section~\ref{sec:scoring}) is separately performed on each rule
graph, and then we use the average of four derived scores as the
final score of a fact.

\textbf{Missing errors.} As the interaction preference and time
conflicts are measured, missing errors can be detected by finding
uninstantiateable nodes with few conflicts in all four rule graphs.
In Section~\ref{sec:time_duration_experiment}, we experimentally
analyze the effectiveness of this strategy.

\section{Experiments}

We conduct extensive experiments on five real-world TKGs and our
experiments aim to answer the following research questions:
\begin{compactitem}
    \item \textbf{RQ1.} How well is \textsc{AnoT} able to detect anomalies?
    \item \textbf{RQ2.} How does each component of \textsc{AnoT} contribute to its performance and is \textsc{AnoT} robust to the hyper-parameters?
    \item \textbf{RQ3.} Is \textsc{AnoT} efficient in detecting anomalies?
    \item \textbf{RQ4.} Is the detection of \textsc{AnoT} interpretable?
    \item \textbf{RQ5.} Can \textsc{AnoT} generalize to the time-duration-based TKGs?
\end{compactitem}

\subsection{Datasets}

Real-world TKGs used in our experiments are shown in
Table~\ref{tab:datasets}. $N_c$, $N_t$, and $N_m$ are the number of
conceptual, time, and missing errors in each dataset. For each TKG,
we use knowledge in the former 60\% timestamps to construct the
model and the latter 40\% for evaluation (10\% for validation and
30\% for testing). We follow previous work
\cite{DBLP:journals/pvldb/FangFGFH23} to inject synthetic anomalies
by randomly perturbing valid knowledge. For each kind of anomaly
(i.e., conceptual, time, and missing errors), we randomly perturb
15\% valid knowledge as the anomalies. For conceptual errors, each
sampled valid knowledge $(s, r, o, t)$ is randomly perturbed as $(s,
r, o', t) \notin \mathcal{F}$ or $(s, r', o, t) \notin \mathcal{F}$,
where $o' \in \mathcal{E}$ and $r' \in \mathcal{R}$. For time
errors, each sampled valid knowledge $(s, r, o, t)$ is randomly
perturbed as $(s, r, o, t')$ where $t \in \mathcal{T}$ and $t'
\notin \mathcal{T}$. We keep a large span between $t$ and $t'$ to
avoid false anomalies. For missing errors, we directly delete the
sampled valid knowledge from TKG to simulate the knowledge missing.
More detailed descriptions of these TKGs are in
\cite{DBLP:conf/sigir/LiJLGGSWC21}. Note that for the
time-duration-based dataset Wikidata, the conceptual errors and
missing errors are generated similarly. The time errors are
generated by randomly perturbing $t_{start}$ or $t_{end}$. We ensure
each invalid knowledge has $t'_{start} \leq t'_{end}$ to avoid being
meaningless.

\begin{table}[t] \small
\caption{Statistics of datasets.} \centering \scalebox{0.92}{
  \begin{tabular}{l||ccccccc}
    \hline
    \textbf{Dataset} & $\mathcal{|E|}$ & $\mathcal{|R|}$ & $\mathcal{|T|}$ & $|\mathcal{F}|$
    & $N_{c}$ & $N_{t}$ & $N_{m}$\\
    \hline
    \textbf{ICEWS 14} &7,128 &230 &365 &90,730 &2,198 &2,198 &2,198 \\
    \textbf{ICEWS 05-15} &10,488 &251 &4,017 &461,329 &13,682 &13,682 &13,682 \\
    \textbf{YAGO 11k} &9,736 &10 &2,801 &161,540 &6,004 &6,004 &6,004 \\
    \textbf{GDELT} &7,691 &240 &2,975 &3,419,607 &91,418 &91,418 &91,418 \\
    \textbf{Wikidata} &12,554 &24 &2,270 &669,934 &9,096 &9,096 &9,096 \\
    \hline
  \end{tabular}}
  \label{tab:datasets}
\end{table}

\subsection{Experimental Setting}

\textbf{Implementation details.} We implement \textsc{AnoT} with
Python and all the experiments are performed with Intel Xeon E5-2650
v3 CPU @ 2.30G Hz processor and 128 GB RAM. We use the officially
released code of baseline models to perform the experiments and for
each baseline model and \textsc{AnoT}, we tune its hyper-parameters
using a grid search. During the candidate generation, we set the
maximum number of categories of entities $k \in \{1, 3, 5, 10\}$. We
also set the maximum number of candidate rule edges as $50000$ to
avoid redundant generation. During the temporal score generation, we
set the maximum step $K \in \{1, 2, 3, 4\}$, and we set the timespan
restriction $L \in \{10, 100, 1000, 2000\}$. For a fair comparison,
we do not allow \textsc{AnoT} to refresh the rule graph during
evaluation.

\textbf{Evaluation protocols.} We evaluate the model performance by
precision ($P$), $F_{\beta}$ score, and area under PR-curve ($AUC$).
For each method, we select the best hyper-parameter settings with
the best $F_{\beta}$ score on the validation set and use $F_{\beta}$
score to select the best threshold. We set $\beta$ as 0.5 to
emphasize the detection precision.

\textbf{Baselines.} We compare \textsc{AnoT} with both temporal
knowledge graph representation learning models and dynamic graph
anomaly detection models. For TKG representation learning models,
typical models in each category are selected, including TNT
\cite{DBLP:conf/iclr/LacroixOU20}, TELM
\cite{DBLP:conf/naacl/XuCNL21}, DE \cite{DBLP:conf/aaai/GoelKBP20},
TA \cite{DBLP:conf/emnlp/Garcia-DuranDN18}, Timeplex
\cite{DBLP:conf/emnlp/JainRMC20}, and RE-GCN
\cite{DBLP:conf/sigir/LiJLGGSWC21}. For dynamic graph anomaly
detection models, since they are inherently unable to handle rich
semantics in TKGs, we only select three typical models: DynAnom
\cite{DBLP:conf/kdd/GuoZS22}, F-FADE
\cite{DBLP:conf/wsdm/Chang0SASL21}, and TADDY
\cite{DBLP:journals/tkde/LiuPWXWCL23}.

\subsection{RQ1: Overall Evaluation}

We report the performance of anomaly detection in
Table~\ref{tab:overall_performance}. Across all anomaly types and
datasets, \textsc{ANoT} achieves the best performance in almost all
the metrics, demonstrating its generality. In particular,
\textsc{ANoT} largely outperforms baseline methods in detecting time
errors (12.2\% on AUC) and missing errors (12.3\% on F-score)
because the baselines neglect the temporal patterns among facts,
whereas \textsc{ANoT} can flexibly infer complex patterns based on
rule edges. RE-GCN can outperform other baselines in detecting
conceptual errors since it integrates the graph structure of TKG.
However, since its performance relies on the richness of the graph
structure and it fails to extract the patterns of interactions,
RE-GCN has a poor performance on sparse dataset (i.e., ICEWS 05-15).

\begin{table}[t]
\caption{Performance comparison of baseline models and \textsc{ANoT}
on inductive anomaly detection. The best results are boldfaced and
the second best results are underlined.} \centering
\resizebox{1\columnwidth}{!}{
\begin{tabular}{c|c|ccc|ccc|ccc|ccc}
\hline &\textbf{Dataset} &\multicolumn{3}{c|}{\textbf{ICEWS 14}}
&\multicolumn{3}{c|}{\textbf{ICEWS 05-15}}
&\multicolumn{3}{c|}{\textbf{YAGO 11k}}
&\multicolumn{3}{c}{\textbf{GDELT}}\\
\hline \textbf{Model} &\textbf{Anomaly} &\emph{Precision}
&\emph{$F_{\beta}$ score}   &\emph{AUC} &\emph{Precision}
&\emph{$F_{\beta}$ score}   &\emph{AUC} &\emph{Precision}
&\emph{$F_{\beta}$ score}   &\emph{AUC}
&\emph{Precision} &\emph{$F_{\beta}$ score}   &\emph{AUC} \\
\hline

&Conceptual errors &0.554 &0.575 &0.867 &0.558 &0.593 &0.861 &0.536 &0.581 &0.877 &0.815 &0.827 &0.863\\
DE
&Time errors &0.273 &0.317 &0.595 &0.309 &0.352 &0.612 &0.335 &0.372 &0.564 &0.283 &0.323 &0.604\\

&Missing errors &0.779 &0.579 &0.690 &0.815 &\underline{0.695} &0.758 &0.924 &0.831 &0.784 &0.630 &0.575 &0.779\\

\hline

&Conceptual errors &0.616 &0.642 &0.887 &0.617 &0.653 &0.890 &0.550 &0.601 &0.884 &0.862 &\textbf{0.843} &0.864\\
TA
&Time errors &0.267 &0.311 &0.579 &0.297 &0.343 &0.512 &0.267 &0.309 &0.589 &0.278 &0.319 &0.562\\

&Missing errors &0.745 &0.537 &0.640 &\underline{0.856} &0.671 &0.709 &0.620 &0.596 &0.720 &0.638 &0.569 &0.764\\

\hline

&Conceptual errors &0.564 &0.588 &0.857 &0.405 &0.445 &0.757 &0.592 &0.636 &0.860 &0.688 &0.615 &0.783\\
Timeplex
&Time errors &0.263 &0.305 &0.513 &0.332 &0.373 &0.547 &0.460 &\underline{0.514} &0.678 &0.330 &0.272 &0.533\\

&Missing errors &0.684 &0.549 &0.682 &0.435 &0.450 &0.608 &0.962 &\underline{0.905} &0.797 &0.433 &0.512 &0.660\\

\hline

&Conceptual errors &0.660 &0.687 &0.904 &0.439 &0.471 &0.780 &0.689 &0.723 &0.917 &0.806 &0.829 &0.855\\
TNT
&Time errors &0.365 &0.409 &0.502 &0.372 &0.407 &0.556 &0.442 &0.490 &0.671 &0.402 &0.430 &0.526\\

&Missing errors &0.610 &0.544 &0.711 &0.632 &0.586 &0.701 &0.955 &0.902 &0.828 &0.635 &0.587 &0.782\\

\hline

&Conceptual errors &0.692 &0.702 &\underline{0.906} &0.509 &0.592 &0.793 &0.659 &0.696 &0.897 &0.866 &0.824 &\underline{0.882}\\
TELM
&Time errors &0.372 &0.416 &0.522 &0.391 &0.421 &0.607 &0.439 &0.489 &0.662 &0.389 &0.417 &0.512\\

&Missing errors &\underline{0.711} &0.562 &0.723 &0.693 &0.611 &0.749 &\underline{0.968} &0.899 &0.815 &0.638 &0.593 &\underline{0.785}\\

\hline

&Conceptual errors &\underline{0.733} &\underline{0.731} &0.901 &\underline{0.737} &\underline{0.712} &\underline{0.920} &\underline{0.817} &\underline{0.825} &\underline{0.934} &\underline{0.870} &0.802 &0.876\\
RE-GCN
&Time errors &0.357 &0.398 &0.617 &0.334 &0.370 &0.667 &0.410 &0.432 &0.687 &0.365 &0.395 &0.628\\

&Missing errors &0.543 &\underline{0.584} &\underline{0.724} &0.763 &0.674 &\underline{0.809} &0.723 &0.724 &\underline{0.840} &\underline{0.674} &\underline{0.629} &0.739\\

\hline \hline

&Conceptual errors &0.565 &0.597 &0.757 &0.516 &0.559 &0.732 &0.759 &0.798 &0.803 &0.621 &0.670 &0.773\\
DynAnom
&Time errors &\underline{0.603} &\underline{0.642} &\underline{0.751} &\underline{0.537} &\textbf{0.528} &\underline{0.723} &0.417 &0.460 &0.682 &\underline{0.620} &\underline{0.669} &\underline{0.771}\\

&Missing errors &0.571 &0.515 &0.652 &0.519 &0.571 &0.679 &0.861 &0.886 &0.831 &0.619 &0.625 &0.728\\

\hline

&Conceptual errors &0.496 &0.544 &0.627 &0.348 &0.378 &0.536 &0.346 &0.397 &0.509 &0.338 &0.342 &0.584\\
F-FADE
&Time errors &0.490 &0.536 &0.615 &0.361 &0.383 &0.514 &0.452 &0.483 &0.502 &0.359 &0.378 &0.551\\

&Missing errors &0.415 &0.465 &0.594 &0.450 &0.479 &0.509 &0.559 &0.613 &0.720 &0.546 &0.557 &0.601\\

\hline

&Conceptual errors &0.329 &0.316 &0.508 &0.313 &0.336 &0.527 &0.493 &0.511 &0.620 &0.370 &0.385 &0.614\\
TADDY
&Time errors &0.517 &0.569 &0.653 &0.344 &0.369 &0.502 &\underline{0.479} &0.496 &\underline{0.691} &0.386 &0.402 &0.593\\

&Missing errors &0.534 &0.572 &0.609 &0.498 &0.514 &0.547 &0.657 &0.683 &0.769 &0.537 &0.543 &0.595\\

\hline \hline

&Conceptual errors &\textbf{0.789} &\textbf{0.792} &\textbf{0.921} &\textbf{0.847} &\textbf{0.815} &\textbf{0.933} &\textbf{0.829} &\textbf{0.841} &\textbf{0.952} &\textbf{0.936} &\underline{0.835} &\textbf{0.887}\\
\textsc{AnoT} (ours)
&Time errors &\textbf{0.639} &\textbf{0.661} &\textbf{0.825} &\textbf{0.601} &\underline{0.526} &\textbf{0.729} &\textbf{0.524} &\textbf{0.579} &\textbf{0.863} &\textbf{0.710} &\textbf{0.786} &\textbf{0.875}\\

&Missing errors &\textbf{0.822} &\textbf{0.705} &\textbf{0.730} &\textbf{0.894} &\textbf{0.797} &\textbf{0.834} &\textbf{0.988} &\textbf{0.933} &\textbf{0.867} &\textbf{0.683} &\textbf{0.696} &\textbf{0.841}\\
\hline
\end{tabular}}
\label{tab:overall_performance}
\end{table}

\begin{figure}[t]
\centering \subfigure[ICEWS
14]{\includegraphics[width=0.24\linewidth]{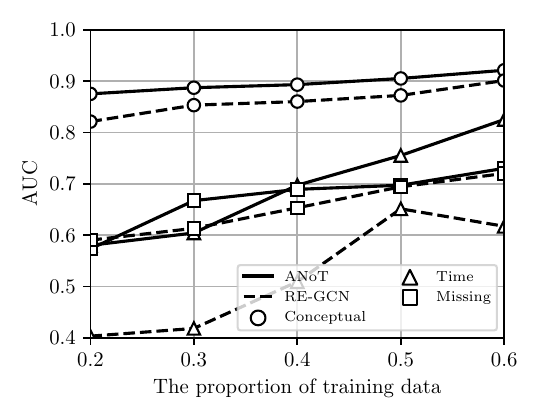}}
\subfigure[ICEWS
05-15]{\includegraphics[width=0.24\linewidth]{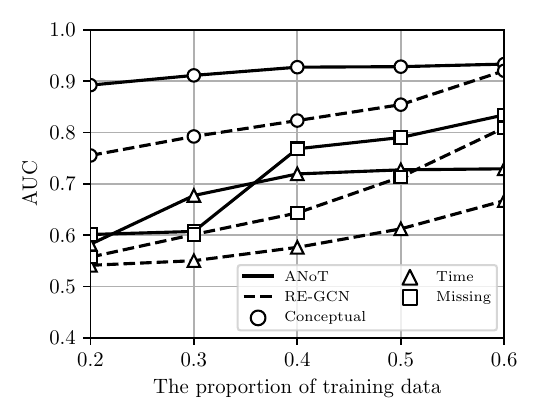}}
\subfigure[YAGO
11k]{\includegraphics[width=0.24\linewidth]{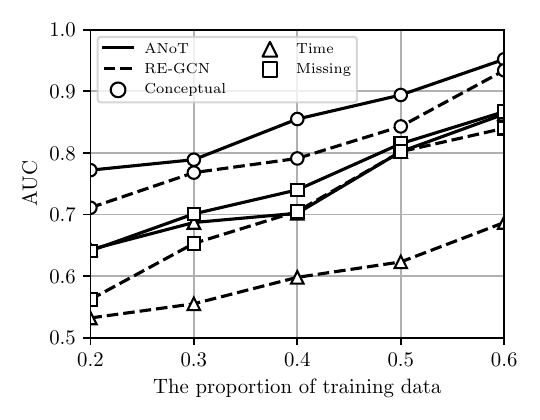}}
\subfigure[GDELT]{\includegraphics[width=0.24\linewidth]{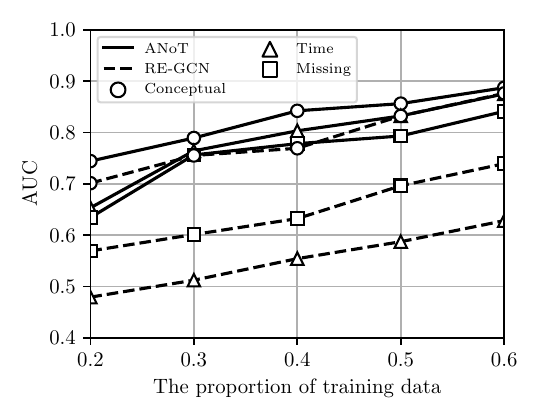}}
\caption{Performance of \textsc{ANoT} and RE-GCN when different
proportions of offline preserved knowledge are used to construct the
optimal rule graph. Concept, time, and missing respectively refer to
conceptual errors, time errors, and missing errors.}
\label{fig:train_proportion}
\end{figure}

To analyze the stability of different methods, in
Figure~\ref{fig:train_proportion} we report detection results of
\textsc{ANoT} and the most powerful baseline method RE-GCN when they
are trained by different proportions of data. \textsc{ANoT} can
outperform RE-GCN in different training proportions. Even though
only 20\% data are used to construct the model, \textsc{ANoT} can
still achieve remarkable detection AUC, which shows its stability.
This result also gives evidence for the effectiveness of our updater
module in adapting the rule graph with new knowledge.

\begin{table}[t]
\caption{Performance comparison of \textsc{ANoT} and its variants.}
\centering \resizebox{1\columnwidth}{!}{
\begin{tabular}{c|c|ccc|ccc|ccc|ccc}
\hline &\textbf{Dataset} &\multicolumn{3}{c|}{\textbf{ICEWS 14}}
&\multicolumn{3}{c|}{\textbf{ICEWS 05-15}}
&\multicolumn{3}{c|}{\textbf{YAGO 11k}}
&\multicolumn{3}{c}{\textbf{GDELT}}\\
\hline \textbf{Variants} &\textbf{Anomaly} &\emph{Precision}
&\emph{$F_{\beta}$ score}   &\emph{AUC} &\emph{Precision}
&\emph{$F_{\beta}$ score}   &\emph{AUC} &\emph{Precision}
&\emph{$F_{\beta}$ score}   &\emph{AUC}
&\emph{Precision} &\emph{$F_{\beta}$ score}   &\emph{AUC} \\

\hline
&Conceptual errors &0.657 &0.680 &0.886 &0.714 &0.736 &0.875 &0.798 &0.807 &0.926 &0.873 &0.796 &0.842\\
Remove category aggregations
&Time errors &0.618 &0.602 &0.758 &0.577 &0.489 &0.676 &0.494 &0.541 &0.846 &0.688 &0.741 &0.833 \\

&Missing errors &0.784 &0.667 &0.681 &0.856 &0.743 &0.801 &0.965 &0.903 &0.842 &0.656 &0.618 &0.799\\

\hline
&Conceptual errors &0.784 &0.781 &0.912 &0.844 &0.810 &0.927 &0.805 &0.815 &0.944 &0.901 &0.817 &0.851\\
Remove updater module
&Time errors &0.551 &0.585 &0.796 &0.530 &0.519 &0.697 &0.438 &0.494 &0.807 &0.670 &0.712 &0.812\\

&Missing errors &0.801 &0.667 &0.689 &0.875 &0.778 &0.807 &0.951 &0.891 &0.840 &0.644 &0.678 &0.809\\

\hline Remove triadic occurring rule edges
&Time errors &0.614 &0.646 &0.812 &0.572 &0.498 &0.685 &0.495 &0.552 &0.846 &0.687 &0.752 &0.844\\

&Missing errors &0.789 &0.693 &0.718 &0.876 &0.792 &0.816 &0.965 &0.912 &0.852 &0.665 &0.635 &0.806\\

\hline

Remove recursive strategy
&Time errors &0.618 &0.634 &0.815 &0.581 &0.522 &0.703 &0.494 &0.550 &0.831 &0.698 &0.761 &0.854\\

&Missing errors &0.792 &0.692 &0.714 &0.882 &0.786 &0.814 &0.962 &0.915 &0.850 &0.670 &0.679 &0.818\\

\hline \hline

&Conceptual errors &0.784 &0.781 &0.912 &0.844 &0.810 &0.927 &0.805 &0.815 &0.944 &0.901 &0.817 &0.851\\
Ranking rules and rule edges only by $|\mathcal{A}^{\mathcal{G}}|
\downarrow$
&Time errors &0.551 &0.585 &0.796 &0.530 &0.519 &0.697 &0.438 &0.494 &0.807 &0.670 &0.712 &0.812\\

&Missing errors &0.801 &0.667 &0.689 &0.875 &0.778 &0.807 &0.951 &0.891 &0.840 &0.644 &0.678 &0.809\\

\hline

&Conceptual errors &0.781 &0.769 &0.901 &0.838 &0.809 &0.931 &0.814 &0.832 &0.941 &0.915 &0.826 &0.877\\
Replace $|\mathcal{A}_v^{\mathcal{G}}|$ as 1 when deriving scores
&Time errors &0.493 &0.448 &0.603 &0.332 &0.375 &0.646 &0.478 &0.529 &0.810 &0.584 &0.632 &0.701\\

&Missing errors  &0.820 &0.584 &0.693 &0.859 &0.732 &0.798 &0.976 &0.873 &0.857 &0.636 &0.642 &0.788\\

\hline \hline

&Conceptual errors &\textbf{0.789} &\textbf{0.792} &\textbf{0.921} &\textbf{0.847} &\textbf{0.815} &\textbf{0.933} &\textbf{0.829} &\textbf{0.841} &\textbf{0.952} &\textbf{0.936} &\textbf{0.835} &\textbf{0.887}\\
Original
&Time errors &\textbf{0.639} &\textbf{0.661} &\textbf{0.825} &\textbf{0.601} &\textbf{0.526} &\textbf{0.729} &\textbf{0.524} &\textbf{0.579} &\textbf{0.863} &\textbf{0.710} &\textbf{0.786} &\textbf{0.875}\\

&Missing errors &\textbf{0.822} &\textbf{0.705} &\textbf{0.730} &\textbf{0.894} &\textbf{0.797} &\textbf{0.834} &\textbf{0.988} &\textbf{0.933} &\textbf{0.867} &\textbf{0.683} &\textbf{0.696} &\textbf{0.841}\\
\hline
\end{tabular}}
\label{tab:ablation_study}
\end{table}

\subsection{RQ2: Effect of Each Component}

\begin{figure}[t]
\centering \subfigure[ICEWS
14]{\includegraphics[width=0.48\linewidth]{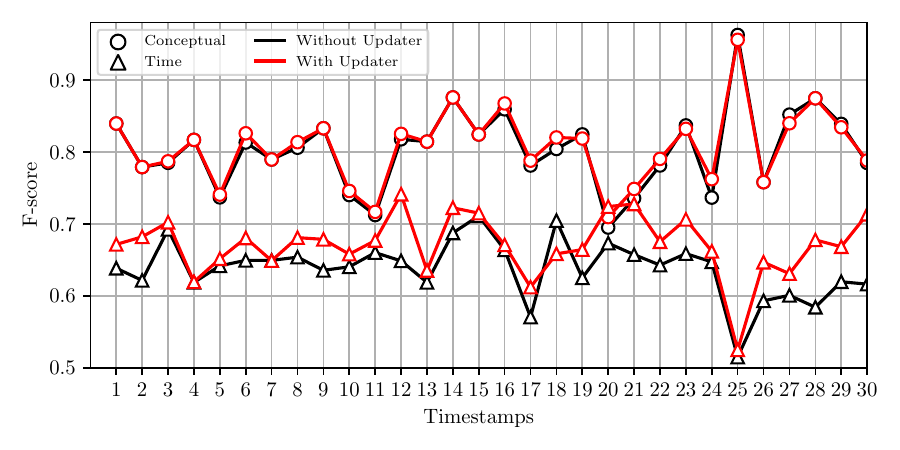}}
\subfigure[GDELT]{\includegraphics[width=0.48\linewidth]{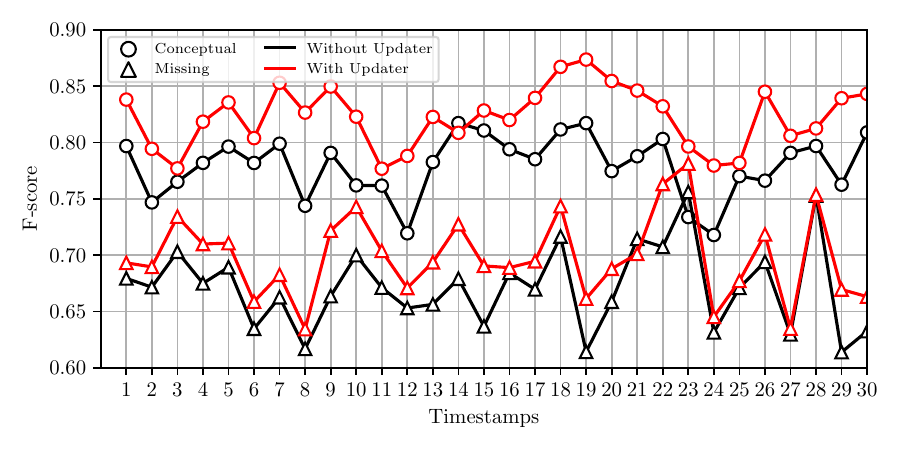}}
\caption{Inductive detection performance of \textsc{ANoT} across
different timestamps on the ICEWS 14 and GDELT datasets.}
\label{fig:long_time_F_score}
\end{figure}

\textbf{Ablation study.} As shown in Table~\ref{tab:ablation_study},
removing category aggregations will result in a degradation in
performance. This is because first, small relation combinations
cannot reflect fine-grained categories; furthermore, aggregation
strategies can summarize redundant categories and thus help to
alleviate the effect of data noises. Removing the updater module
will result in the model failing to adapt to the new knowledge, thus
degrading the performance, especially in detecting time errors. This
is because temporal patterns change more frequently and thus require
timely updating. Figure~\ref{fig:long_time_F_score} shows the
long-time detection performance, particularly for unseen knowledge,
with and without the updater module. We can see that the updater can
constantly enhance performance across different timestamps,
demonstrating its effectiveness in handling new emerging knowledge
and patterns. Since triadic rule edges and recursive strategies only
affect the temporal scoring, only results of time errors and missing
errors are reported. We can see that they both contribute to the
detection performance, demonstrating the effectiveness of triadic
rule edges in describing knowledge-occurring patterns and recursive
strategies in supporting more stable scoring.

\textbf{Comparison with variants.} We further analyze the
effectiveness of our ranking strategy. As shown in
Table~\ref{tab:ablation_study}, when we rank the atomic rules and
rule edges only by the number of correct assertions
$|\mathcal{A}^{\mathcal{G}}|$, the performance of all anomaly types
degrades. This is because the rules and rule edges with larger
$|\mathcal{A}^{\mathcal{G}}|$ are not necessarily helpful in
improving the expressiveness of the model and thus may introduce
low-expressive rules and edges. When we only use the number of
mapped rules and rule edges to derive scores, the performance
degrades especially for the time errors. This shows the
effectiveness of considering different expression powers.

\begin{figure}[t]
\centering \subfigure[ICEWS
14]{\includegraphics[width=0.24\linewidth]{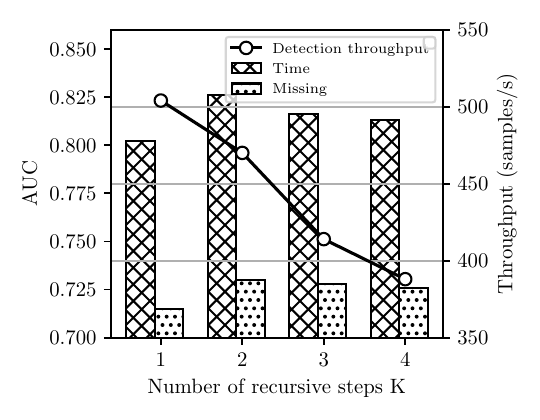}}
\subfigure[ICEWS
05-15]{\includegraphics[width=0.24\linewidth]{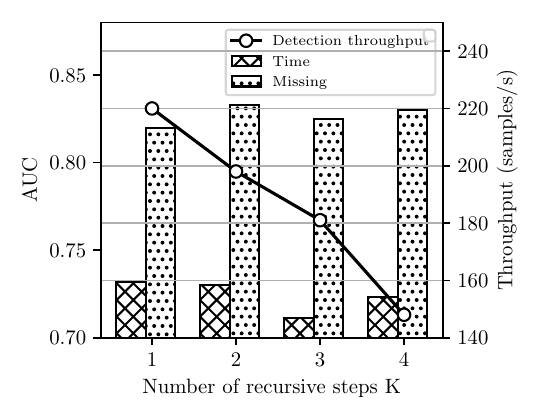}}
\subfigure[YAGO
11k]{\includegraphics[width=0.24\linewidth]{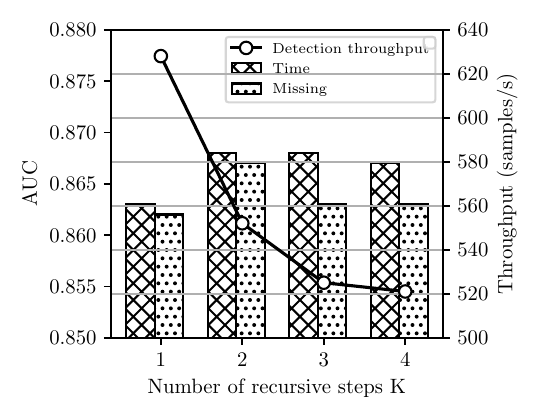}}
\subfigure[GDELT]{\includegraphics[width=0.24\linewidth]{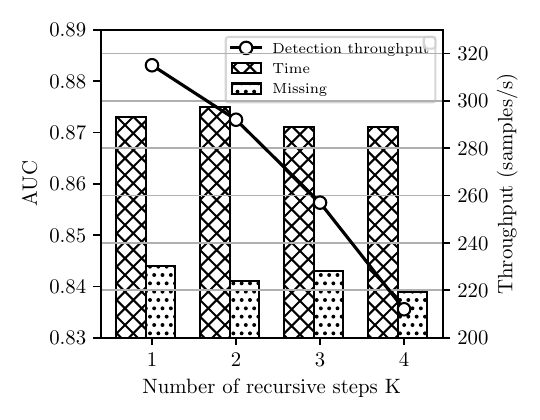}}
\caption{Performance and detection throughput of \textsc{ANoT} under
the different settings of the number of recursive steps $K$.}
\label{fig:steps}
\end{figure}

\textbf{Effects of the number of recursive steps $K$.} As shown in
Figure~\ref{fig:steps}, when $K$ increases from 1 to 2, the
detection performances on all four datasets improve, which
demonstrates the effectiveness of our recursive strategy in
achieving more accurate scoring. However, when $K$ continues to
increase, the performance will gradually degrade. This is because
the more steps required by a reachable node, the more uncertain it
can be evidence for new knowledge. Thus, a too large $K$ will bring
much noise during scoring.

\begin{figure}[t]
\centering \subfigure[ICEWS
14]{\includegraphics[width=0.24\linewidth]{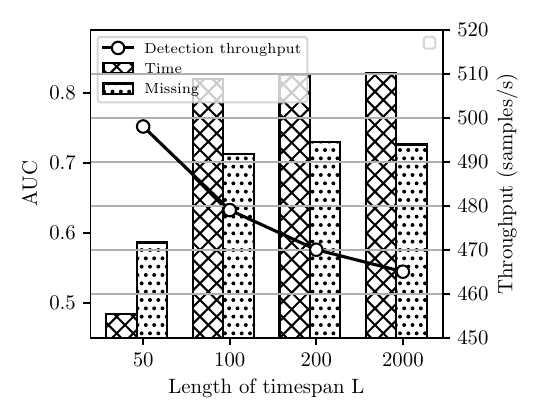}}
\subfigure[ICEWS
05-15]{\includegraphics[width=0.24\linewidth]{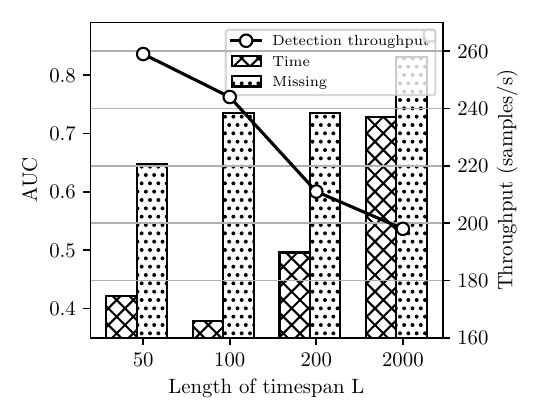}}
\subfigure[YAGO
11k]{\includegraphics[width=0.24\linewidth]{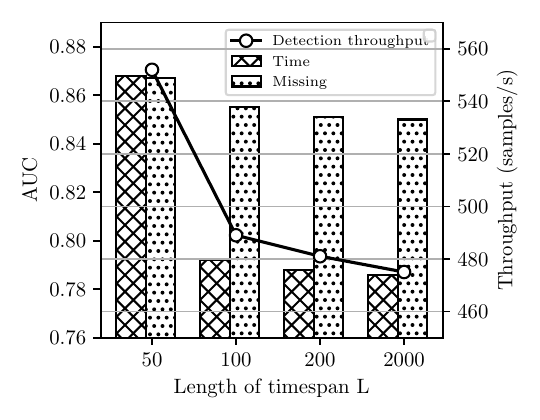}}
\subfigure[GDELT]{\includegraphics[width=0.24\linewidth]{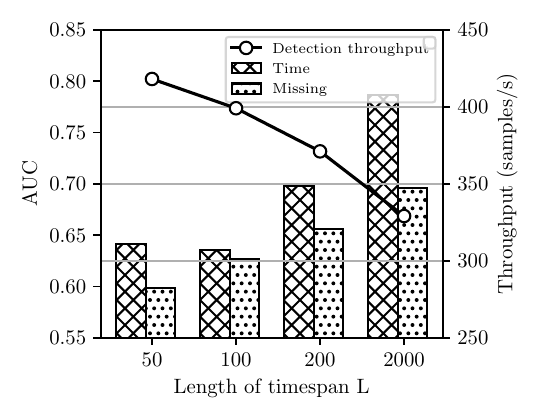}}
\caption{Performance and detection throughput of \textsc{ANoT} under
the different settings of the timespan length $L$.}
\label{fig:spans}
\end{figure}

\textbf{Effects of the length of timespan $L$.} As shown in
Figure~\ref{fig:spans}, different datasets require different $L$ to
get the best performance. We notice that the best $L$ is largely
related to the size of each dataset (e.g., 200 for ICEWS 14 and 2000
for ICEWS 05-15). This may be because a larger TKG requires a larger
$L$ to extract long-range patterns. The YAGO 11k dataset requires a
small $L$ since its granularity is a month but other datasets are
day or minute.

\begin{figure}[t]
\centering \subfigure[ICEWS
14]{\includegraphics[width=0.24\linewidth]{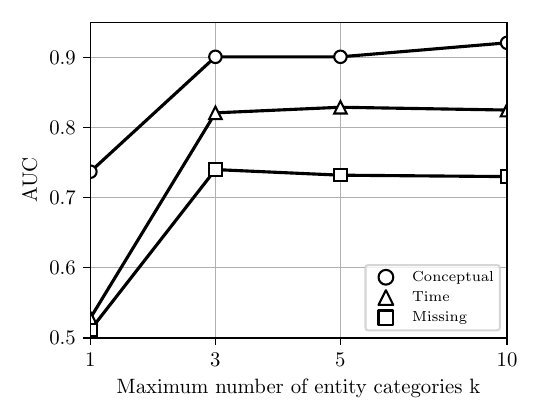}}
\subfigure[ICEWS
05-15]{\includegraphics[width=0.24\linewidth]{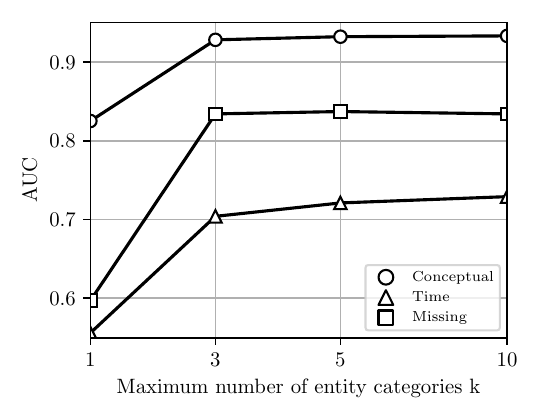}}
\subfigure[YAGO
11k]{\includegraphics[width=0.24\linewidth]{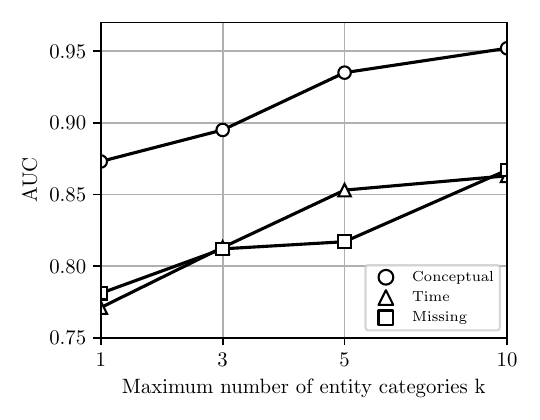}}
\subfigure[GDELT]{\includegraphics[width=0.24\linewidth]{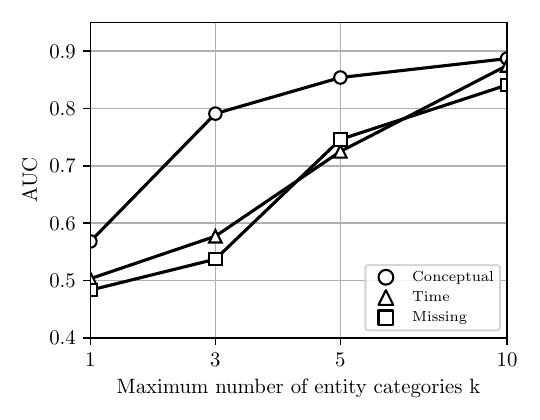}}
\caption{Performance of \textsc{ANoT} under the different settings
of the number of entity categories $k$.} \label{fig:cat_num}
\end{figure}

\textbf{Effects of the number of entity categories $k$.} As shown in
Figure~\ref{fig:cat_num}, two ICEWS datasets get the nearly best
performance when $k$ is 3, and keep stable when $k$ continues to be
larger. This demonstrates that our category function can effectively
extract the properties shared by different entities. The YAGO 11k
dataset has a smaller number of relations which limits the
expression power of each relation combination and thus requires more
categories.

\subsection{RQ3: Efficiency Analysis}

\textbf{Detection efficiency.} As shown in Figure~\ref{fig:steps},
we analyze how the recursive steps $K$ affect the detection
efficiency. With $K$ increases, the throughput of \textsc{ANoT}
gradually decreases, but we can see that the best performance of
\textsc{ANoT} does not rely on a large $K$. Therefore, a small $K$
can already achieve remarkable performance and acceptable
efficiency. Furthermore, When $K$ increases, the reachable nodes
increase exponentially but our throughput only decreases linearly,
which also shows the effectiveness of our strategies.

In Figure~\ref{fig:spans}, we analyze how the timespan $L$ affects
the detection efficiency. We find that even though $L$ increases
largely, the throughput only decreases linearly. This is because our
ranking and selection strategies can avoid too many redundant rule
edges being added to the rule graph, and thus reduce the searching
time.

\begin{table}[t]
\caption{Model building time, the sizes of the obtained optimal rule
graph, and the proportions of explained facts under different
settings of category number $K$.} \centering
\resizebox{1\columnwidth}{!}{
\begin{tabular}{c|cccc|cccc|cccc|cccc}
\hline &\multicolumn{4}{c|}{\textbf{ICEWS 14}}
&\multicolumn{4}{c|}{\textbf{ICEWS 05-15}}
&\multicolumn{4}{c|}{\textbf{YAGO 11k}}
&\multicolumn{4}{c}{\textbf{GDELT}}\\
\hline \textbf{Maximum number of entity categories $k$} &\emph{1}
&\emph{3} &\emph{5} &\emph{10} &\emph{1} &\emph{3} &\emph{5}
&\emph{10} &\emph{1} &\emph{3} &\emph{5} &\emph{10}
&\emph{1} &\emph{3} &\emph{5} &\emph{10} \\

\hline
Building time &49s &584s &629s &660s &773s &1,239s &1,577s &1,814s &44s &50s &65s &84s &1,160s &1,585s &2,081s &2,314s\\

Number of rule edges &97 &6,349 &5,428 &5,064 &3,751 &9,688 &11,053 &11,384 &92 &76 &66 &63 &87 &20,223 &27,847 &30,106 \\

Proportion of explained facts &0.132 &0.766 &0.767 &0.767 &0.165 &0.788 &0.853 &0.874 &0.716 &0.909 &0.908 &0.907 &0.170 &0.896 &0.894 &0.894\\
\hline
\end{tabular}}
\label{tab:efficiency}
\end{table}

\textbf{Time consumption of building model.} As shown in
Table~\ref{tab:efficiency}, the time consumption of building the
optimal rule graph increases sub-linearly when the number of entity
categories $k$ increases, which is acceptable in practice. The
consumption of the YAGO 11k dataset is extremely small since it
contains only a few relation categories, therefore the option spaces
of the candidate rules and rule edges are smaller than other
datasets. The size of the GDELT dataset is much larger than other
datasets but our \textsc{ANoT} can still handle it within one hour
(baselines need over 100 epochs to train while each epoch needs
nearly 3 minutes), showing our efficiency.

\textbf{Size of the optimal rule graph.} We further analyze the size
of the obtained optimal rule graph in Table~\ref{tab:efficiency}. As
the number of entity categories $k$ continuously increases, the
number of rule edges becomes stable or even slightly decreases,
meaning that the construction of the optimal rule graph is robust to
the entity categories. The decrease may be because a larger number
of entity categories can expand the number of candidate rules and
rule edges, and thus help \textsc{ANoT} to find more powerful
patterns. We further find that as $k$ increases, the proportion of
facts that can be explained gradually increases, which means
improvements in the expression power of our rule graph. Combined
with the results in Figure~\ref{fig:cat_num}, the case with a high
proportion of explained facts also has a high AUC in detection,
showing the effectiveness of our selection strategies.

\begin{table}[t]
\caption{Examples of the obtained entity categories and the entities
that are assigned as these categories.} \centering
\resizebox{1\columnwidth}{!}{
\begin{tabular}{c|c}
\hline \multicolumn{1}{c|}{\textbf{Entity category (relation
combinations)}}& \multicolumn{1}{c}{\textbf{Described entities}}
\\ \hline
host a visit | Express intent to provide military aid | Make
statement | Express intent to change leadership & 1. Xi Jinping, 2.
Barack Obama, 3. Kim Jong-Un
\\ \hline
Demand economic aid | Threaten with military force | Return or
release person(s) & 1. Naxalites group, 2. Rebel Group (Abu Sayyaf),
3. Combatant (Djibouti)
\\ \hline
Died in | Was born in & 1. China, 2. Japan, 3. United States 4.
France
\\ \hline
Has won prize & 1. Nobel Peace Prize, 2. Fields Medal 3. Lasker
Award
\\ \hline
 Reduce or break diplomatic relations | Bring a lawsuit against & 1. Rights Activist (Ukraine), 2. Shamsul Islam Khan 3. Center for Reproductive Rights 4. Democratic Labor Party
\\ \hline
Was born in | Created | Graduated from & 1. Harry Weese, 2. I. M.
Pei 3. Whit Stillman 4. Robert Ardrey
\\ \hline
\end{tabular}}
\label{tab:cat_interpretability}
\end{table}

\subsection{RQ4: Interpretability Analysis}

\textbf{Category interpretability.} We select some representative
entity categories and report their corresponding relation
combinations and the entities that are assigned as these categories
in Table~\ref{tab:cat_interpretability}. We can see that relations
contained in one relation combination have related semantics, e.g.,
`Express intent to provide military aid' and `Make statement' are
both political behaviors of leaders. Thus, relation combinations can
imply entity categories. For example, `Was born in' and `Created'
may imply artists, and `Express intent to provide military aid' and
`Make statement' may imply presidents. The described entities of
these categories also give evidence, e.g.,, `Barack Obama' and `Kim
Joung-Un' are both presidents, and `Harry Weese' and `I. M. Pei' are
both architects. These examples show the interpretability of our
framework at the atomic rule level.

\begin{table}[t]
\caption{Examples of rule edges in the obtained optimal rule graph.}
\centering \resizebox{1\columnwidth}{!}{
\begin{tabular}{c}
\hline \multicolumn{1}{c}{\textbf{Entity category (relation
combinations)}}
\\ \hline
(PERSON, Was born in, COUNTRY) $\rightarrow$ (PERSON, Died in,
COUNTRY)
\\ \hline
(PERSON, Created, PRODUCTS) $\rightarrow$ (PERSON, Owns, PRODUCTS)
\\ \hline
(COUNTRY, Host a visit, PERSON) $\rightarrow$ (PERSON, Make a visit,
COUNTRY)
\\ \hline
(ORGANIZATION A, Cooperate, ORGANIZATION B) $\rightarrow$
(ORGANIZATION A, Consult, ORGANIZATION B)
\\ \hline \hline
(COUNTRY A, Accuse, PEOPLE), (COUNTRY A, Provide military aid,
COUNTRY B) $\rightarrow$ (COUNTRY B, Accuse, PEOPLE)
\\ \hline
(PERSON A, Was born in, COUNTRY), (PERSON A, Is married to, PERSON
B)  $\rightarrow$ (PERSON B, Was died in, COUNTRY)
\\ \hline
(COUNTRY A, Investigate, COUNTRY B), (COUNTRY C, Criticize or
denounce, COUNTRY B) $\rightarrow$ (COUNTRY A, Express intent to
cooperate, COUNTRY C)
\\ \hline
(COUNTRY A, Engage in negotiation, COUNTRY B), (COUNTRY C, Halt
negotiations, COUNTRY B) $\rightarrow$ (COUNTRY A, Express intent to
meet or negotiate, COUNTRY C)
\\ \hline
\end{tabular}}
\label{tab:edge_interpretability}
\end{table}

\textbf{Rule edge interpretability.} In
Table~\ref{tab:edge_interpretability} we further select some
representative rule edges to show the interpretability of our
framework at the rule edge level. The chain-based rule edges extract
some direct relevance between facts, such as `Create' $\rightarrow$
`Owns', while the triadic-based rule edges can extract more complex
relevance, such as `Accuse', `Provide military aid' $\rightarrow$
`Accuse' shows that countries with cooperation tend to have the same
position. Based on these interpretable rules and rule edges, our
\textsc{ANoT} can generate a set of human-readable prompts to
describe what patterns a new knowledge comply and what patterns it
violates.

\subsection{RQ5: Generalization Analysis}
\label{sec:time_duration_experiment}

\textbf{Detection accuracy.} Here we analyze the generalization
ability of \textsc{ANoT} to the time-duration-based TKGs. We employ
the most popular time-duration TKG benchmark Wikidata
\cite{DBLP:conf/sigir/LiJLGGSWC21} for evaluation. As shown in
Table~\ref{tab:Wikidata_performance}, \textsc{ANoT} can still
outperform existing TKG embedding models, especially for the time
errors, demonstrating its generalization ability, and the updater
module can help our solution to adapt to time-duration-based pattern
changes.

\begin{figure}[t]
\centering
\subfigure[]{\includegraphics[width=0.42\linewidth]{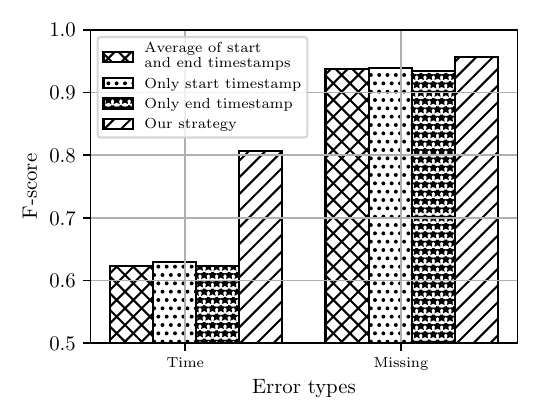}}
\subfigure[]{\includegraphics[width=0.42\linewidth]{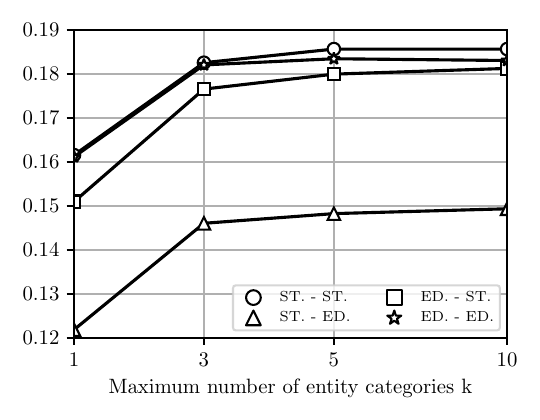}}
\caption{(a) Performance of different adaption strategies. (b)
Proportions of facts explained by different rule graphs.}
\label{fig:Wikidata_analysis}
\end{figure}

\textbf{Effectiveness of adaption strategy.} As illustrated in
Figure~\ref{fig:Wikidata_analysis}(a), we can see that our strategy
outperforms other simple strategies that can transfer time duration
to timestamps, showing the effectiveness of our proposed four types
of rule graphs in capturing various patterns in time-duration-based
TKGs.

\textbf{Effectiveness of different rule graphs.}
Figure~\ref{fig:Wikidata_analysis}(b) shows how four types of rule
graphs contribute to the performance. We can see that all four types
of rule graphs can describe a unique part of time-duration
knowledge, showing their necessities. As the number of entity
categories increases, each type of rule graph can describe more
knowledge.

\begin{table}[t]
\caption{$F_{\beta}$ score of \textsc{ANoT} and typical TKG
embedding models on the time-duration-based TKG dataset Wikidata.}
\centering \scalebox{0.68}{
\begin{tabular}{c|ccc}
\hline \textbf{Model}
&Conceptual errors &Time errors   &Missing errors\\
\hline
DE &0.849 &0.560 &0.869\\

TA &0.859 &0.530 &0.701\\

Timeplex &0.866 &0.668 &0.898\\

TNT &0.806 &0.586 &0.879\\

TELM &0.851 &0.641 &0.885\\

RE-GCN &0.858 &\underline{0.697} &0.903\\

\hline \hline
ANoT (without updater) &\underline{0.961} &0.687 &\underline{0.951}\\

ANoT &\textbf{0.967} &\textbf{0.806} &\textbf{0.956}\\
\hline
\end{tabular}}
\label{tab:Wikidata_performance}
\end{table}

\section{Conclusion}

In this paper, we make the first attempt at strategies to summarize
a temporal knowledge graph and first explore how to inductively
detect anomalies in TKG. We propose a novel rule graph to map a TKG
as a set of human-readable rules and rule edges. The rule graph
allows us to flexibly infer complex patterns. Based on the rule
graph, we propose an \textsc{ANoT} framework, which can efficiently
detect anomaly knowledge by traversing the rule graph, and
effectively adapt the rule graph to new knowledge. Extensive
experimental results demonstrate the superiority of \textsc{ANoT} in
accuracy, robustness, efficiency, and interpretability. In our
future works, integrating our rule graph with graph learning methods
is an interesting direction.

\begin{acks}
This work is supported by the National Natural Science Foundation of
China (No. 62276047) and Shenzhen Science and Technology Program
(No. JCYJ20210324121213037).
\end{acks}

\bibliographystyle{ACM-Reference-Format}
\bibliography{sample-base}

\end{document}